\documentclass[a4paper,fleqn]{cas-dc}

\usepackage[numbers]{natbib}

\usepackage{float}
\usepackage{subfigure}
\usepackage{times}
\usepackage{epsfig}
\usepackage{graphicx}
\usepackage{amsmath}
\usepackage{amssymb}

\usepackage{wrapfig}
\usepackage{graphicx}

\usepackage{caption2}

\begin{document}
%\renewcommand{\figurename}{Fig.} % 加入的代码
%\captionsetup[figure]{labelfont={bf},labelformat={default},labelsep=period,name={Fig.}}

%\let\WriteBookmarks\relax
%\def\floatpagepagefraction{1}
%\def\textpagefraction{.001}
%\shorttitle{Leveraging social media news}
%\shortauthors{CV Radhakrishnan et~al.}

\title [mode = title]{Crowd counting with crowd attention convolutional neural network}                      
%\tnotemark[1,2]

%\tnotetext[1]{This document is the results of the research
%   project funded by the National Science Foundation.}

%\tnotetext[2]{The second title footnote which is a longer text matter
%   to fill through the whole text width and overflow into
%   another line in the footnotes area of the first page.}

%\renewcommand{\Authand}{, }

\author[1, 2]{Jiwei Chen}%[type=editor,
                        %auid=000,bioid=1,
                        %prefix=Sir,
                        %role=Researcher,
                        %orcid=0000-0001-7511-2910]
%\ead{cvr3@sayahna.org}   %邮箱
%\ead[URL]{www.sayahna.org}
\author[3]{Su Wen}
%\author[2]{Su Wn}
\credit{Conceptualization of this study, Methodology, Software}

%\address[1]{Institute of Intelligent Machines, Chinese Academy of Sciences, Hefei, China}
%\address[2]{University of Science and Technology of China, Hefei, China}

\author[1,2]{Zengfu Wang}[orcid=0000-0003-1859-900X]
%\cormark[1]
\ead{Corresponding author:zfwang@ustc.edu.cn}
%\email{zfwang@ustc.edu.cn}
%\Letter{zfwang@ustc.edu.cn}
%\thanks{Corresponding author: email@mail.com}
%\ead[url]{www.cvr.cc, cvr@sayahna.org}
%\author[2,3]{CV Rajagopal}[%
%   role=Co-ordinator,
%   suffix=Jr,
%   ]

%\credit{Data curation, Writing - Original draft preparation}

\address[1]{Institute of Intelligent Machines, Chinese Academy of Sciences, Hefei, China}
\address[2]{Department of Automation, University of Science and Technology of China, Hefei, China}
\address[3]{Faculty of Mechanical Engineering and Automation, Zhejiang Sci-Tech University, Hangzhou, China}

%\author%
%[1,3]
%{Rishi T.}
%\cormark[2]
%\fnmark[1,3]
%\ead{rishi@stmdocs.in}
%\ead[URL]{www.stmdocs.in}

%\address[3]{STM Document Engineering Pvt Ltd., Mepukada,
%    Malayinkil, Trivandrum 695571, India}

%\cortext[cor2]{Corresponding author.}
%\cortext[cor1]{Corresponding author.}
%\cortext[cor2]{Principal corresponding author}
%\fntext[fn1]{This is the first author footnote. but is common to third
%  author as well.}
%\fntext[fn2]{Another author footnote, this is a very long footnote and
 %it should be a really long footnote. But this footnote is not yet
 %sufficiently long enough to make two lines of footnote text.}

%\nonumnote{This note has no numbers. In this work we demonstrate $a_b$
%  the formation Y\_1 of a new type of polariton on the interface
%  between a cuprous oxide slab and a polystyrene micro-sphere placed
%  on the slab.
%  }

\begin{abstract}
Crowd counting is a challenging problem due to the scene complexity and scale variation. Although deep learning has achieved great improvement in crowd counting, scene complexity affects the judgement of these methods and they usually regard some objects as people mistakenly; causing potentially enormous errors in the crowd counting result. To address the problem, we propose a novel end-to-end model called Crowd Attention Convolutional Neural Network (CAT-CNN). Our CAT-CNN can adaptively assess the importance of a human head at each pixel location by automatically encoding a confidence map. With the guidance of the confidence map, the position of human head in estimated density map gets more attention to encode the final density map, which can avoid enormous misjudgements effectively. The crowd count can be obtained by integrating the final density map. To encode a highly refined density map, the total crowd count of each image is classified in a designed classification task and we first explicitly map the prior of the population-level category to feature maps. To verify the efficiency of our proposed method, extensive experiments are conducted on three highly challenging datasets. Results establish the superiority of our method over many state-of-the-art methods. % demonstrate that our method outperforms many state-of-the-art methods. on UCF\_CC\_50 dataset. and reveals 34.24 improvement on the second best method.

\end{abstract}

%\includegraphics{figs/grabs.pdf}

%\item Research highlights item 1
%\item Research highlights item 2
%\item Research highlights item 3

\begin{keywords}

Convolutional neural network  \sep Crowd counting   \sep Confidence map  \sep Density map
\end{keywords}

\maketitle

\section{Introduction}
Crowd counting by computer vision technology plays an important role in safety management \cite{zhang2015cross}, video surveillance \cite{zhan2008crowd}, and urban planning \cite{ryan2009crowd}. The method of crowd counting can be also extended to other applications \cite{marsde2018people}, such as cell counting, animal counting, and vehicle counting. However, due to the severe occlusion, scale variation, and high density in the crowd scene, crowd counting is still a challenging task.

To address these problems, a lot of efforts \cite{sindagi2018survey,li2015crowded} have been done in previous works including detection-based methods \cite{zeng2010robust,li2008estimating,lin2010shape} and regression-based methods \cite{chan2008privacy,chan2009bayesian,idrees2013multi}. Detection-based methods \cite{zeng2010robust,li2008estimating,lin2010shape} usually detect the instances of each person with pre-trained detectors \cite{dalal2005histograms,wang2009hog}. In the sparse crowd scene, they count the crowd accurately, while their accuracies are downgraded in the congested scene. Regression-based methods \cite{chan2008privacy,chan2009bayesian,idrees2013multi} regress the number of the crowd without detecting people. They implement an implicit mapping between low-level features and crowd counts. However, the location information of the crowd is omitted. So that many CNN-based methods with state-of-the-art results \cite{zhang2016single,sam2017switching,sindagi2017generating} are proposed recently. Most of them map the image to a density map that is more robust than the hand-crafted features. The quantity and location of the crowd at each pixel location are recorded in the density map. The crowd count can be obtained by integrating the density map.

%redmon2016you,ren2015faster,\cite{walach2016learning,liu2018facial,zhai2018generative}
%The crowd scene analysis using computer vision technology plays an important role in the crowd's management and security. In recent years, the deep learning has obtained breakthroughs in many fields \cite{walach2016learning,liu2018facial,zhai2018generative}, and many CNN models \cite{zhang2016single,xiong2017spatiotemporal,sindagi2017generating} are proposed in crowd counting. Researchers \cite{sam2017switching,amirgholipour2018ccnn} use the CNN models to get a density map which contains count and spatial information at each pixel, then the density map is integrated to obtain the crowd count results. 

  Although CNN-based methods have achieved significant success in crowd counting, we find an important problem that needs to be solved urgently. Due to the complexity of crowd scenes, CNN-based methods usually mistake some objects as the head of people. As shown in Fig. 1, there are no people inside the red box, however, MCNN \cite{zhang2016single} regards the dense shrubberies as human heads by mistake, which results in enormous errors of crowd counting.

  To address the above problem, we propose a novel end-to-end model called CAT-CNN. An overview of the proposed CAT-CNN is shown in Fig. 2. It contains four modules: Multi-information Handling Module, Confidence Module, Density Map Estimation Module, and Fusion Module. The Multi-information Handling Module is utilized to extract robust features for crowd counting. Motivated by \cite{boominathan2016crowdnet,zhang2016single}, we leverage different convolution kernels to encode the input image at the beginning, then we fuse rich hierarchies from different convolutional layers, which is significant for extracting multi-scale features. In addition, the total crowd count of each image is classified \cite{li2018pdisvpl} in a designed crowd count group classifier. To the best of our knowledge, we first explicitly map the weights of predicted class to feature maps to automatically contribute in encoding a highly refined density map. In the Confidence Module, we classify each pixel to obtain the probability of a human head at each pixel location to encode the confidence map. Unfortunately, the ground-truth confidence map is not provided in present crowd counting datasets. We propose a simple but effective way to obtain the ground-truth confidence map by pasting the ones template on a binary map. The intensive cost of manual labeling is saved. Meanwhile, to address the problem of unbalanced population distribution, we propose the weighted Binary Cross-Entropy Loss (BCELoss) to encode a robust confidence map for population distribution. In the Density Map Estimation Module, the estimated density map is encoded. In the Fusion Module, the estimated density map is multiplied by the pixel-level confidence map. With the guidance of confidence map, the position of human head in the estimated density map gets more attention to encode the final density map and enormous misjudgements are avoided effectively. The final density map is integrated to get the crowd count. These modules work collaboratively to complete the crowd counting task. The training method of them is not as complex as these methods in \cite{zhang2016single,sam2017switching,sindagi2017generating}. They are trained jointly by minimizing their loss functions. They don't need pre-training and need to be trained only once.

  %In this paper, we design a novel model called CAT-CNN. As shown in Fig.1(d), our method can solve the problem better. CAT-CNN can produce the confidence map to guide our network to reduce the effect of complex backgrounds and increase the accuracy of crowd counting. And we first propose classifying each pixel to obtain the possibility of a human head at each pixel location to reduce wrong judgements in crowd counting. As shown in Fig.3, the CAT-CNN contains four modules. These modules work collaboratively to complete the multi-process task, they are trained by minimizing their loss functions simultaneously.
 
   %In this paper, we propose a novel end-to-end CNN to achieve lower count errors. We first propose classifying each pixel to obtain the possibility of a human head at each pixel location to reduce wrong judgements in crowd counting. our model does not need to be pre-trained on other datasets. 
   Our contributions are summarized as follows:

   1. We propose the CAT-CNN that can adaptively assess the importance of a human head at each pixel location to avoid enormous misjudgements in crowd counting.
% With the guidance of the confidence map, the final accurate density map is encoded from the estimated density map.

   2. We design a novel classification model that can take input of arbitrary size for training in crowd counting. And we first explicitly map the prior information of the population-level category of images to feature maps to automatically contribute in encoding a highly refined density map.

   %3. We propose a simple but effective way to obtain the ground-truth confidence map (mask) by pasting the ones template on a binary map. In addition, the weighted Binary Cross Entropy Loss (BCELoss) is proposed to encode a robust confidence map for population distribution.

   3. Our CAT-CNN is a multi-stage and multi-supervision model. Meanwhile, it is robust to scale variations by the novel design in the Multi-information Handling Module. Extensive experiments demonstrate that our method outperforms many state-of-the-art methods on three highly challenging datasets (\cite{idrees2013multi,zhang2016single,zhang2015cross}). %on UCF\_CC\_50 datasets.

\begin{figure}
	\centering
   \includegraphics[width=0.85\linewidth]{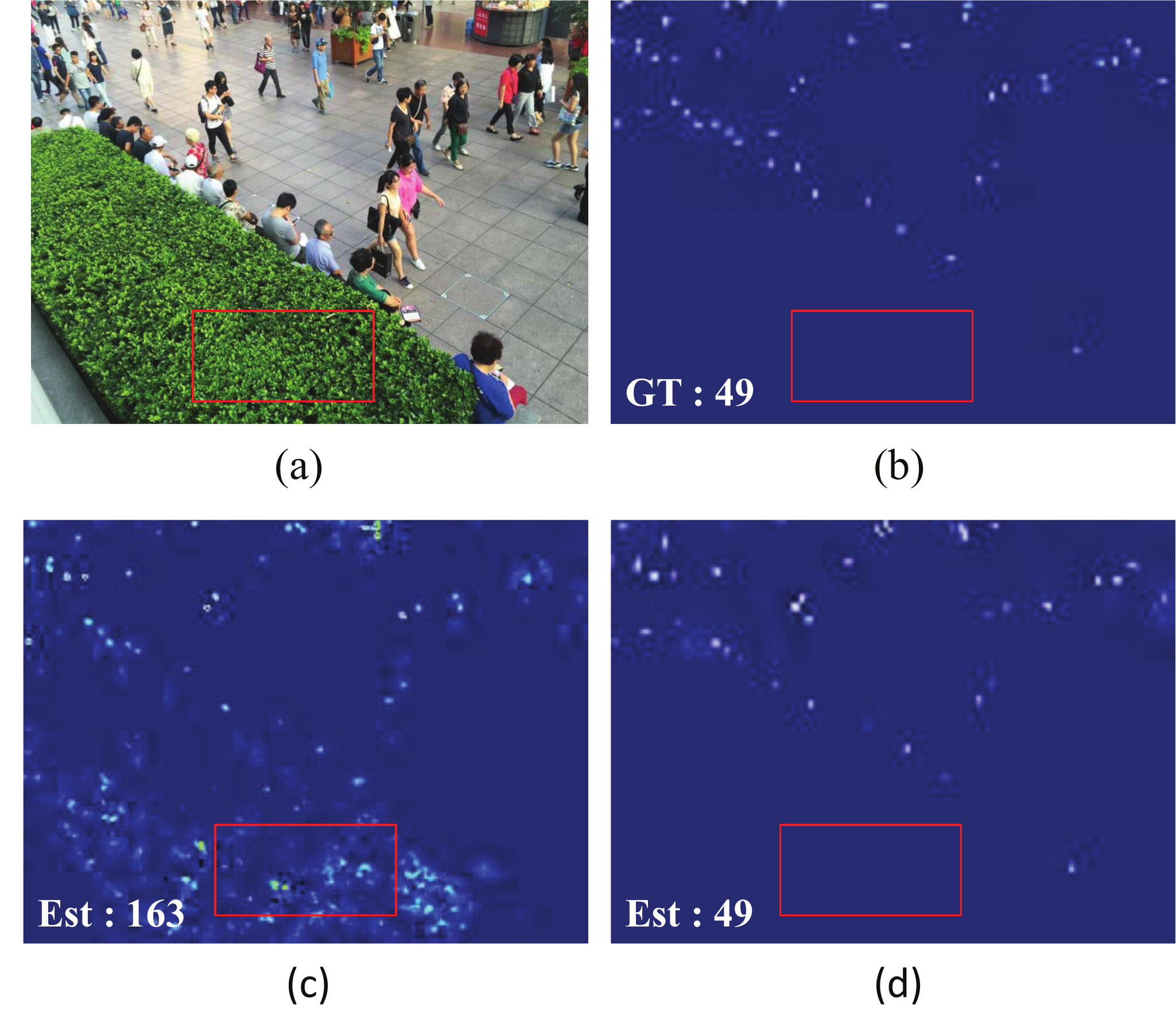}

   \caption{Density estimation results. (a) Input image.      (b) Ground-truth density map. (c) The density map estimated by MCNN. (d) The density map estimated by CAT-CNN. GT represents the ground-truth count. Est represents the estimated count.}
	\label{FIG:1}
\end{figure}

\section{Related work}
In recent years, crowd counting has drawn much attention and various methods have been proposed, especially in deep learning. Next, we will give these methods some introductions. 

%Recently,  \cite{uijlings2013selective} first use region proposal generators to detect the areas where people may be present.

\subsection{Detection-based methods}
Traditional detection-based algorithms such as Haar wavelets \cite{viola2004robust}, HOG \cite{dalal2005histograms}, and LBP  \cite{wang2009hog} occupy an important position in early works. Lin et al. \cite{li2008estimating} employed the Haar wavelet transform to detect the head-like contour. Zeng et al. \cite{zeng2010robust} detected the head-shoulder of people by HOG and LBP. Lin et al. \cite{lin2010shape} proposed a part-template tree model for human detection. The crowd count can be obtained by summing the total number of positive samples in their methods. Detection-based methods perform very well in the sparse crowd. However, when the crowd becomes dense, some people are too small to be detected.

\subsection{Regression-based methods}

Since the accuracy of detection-based method is not very high in the highly congested scene, researchers attempt to use regression-based methods to handle this problem. The regression-based methods learn a mapping between high-level features and crowd counts. The high-level features are extracted from low-level information such as edge information \cite{chan2008privacy}, texture information \cite{marana1998efficacy}, and segmentation information \cite{ ryan2009crowd}, then the crowd count is regressed according to high-level features. Chan et al. \cite{chan2008privacy} proposed the Gaussian regression algorithm to learn a mapping between feature maps and crowd counts. Chan et al. \cite{chan2009bayesian} employed the Poisson regression algorithm to model the crowd count as the Poisson random variable. To decrease crowd counting errors, Idrees et al. \cite{idrees2013multi} utilized some outstanding features such as people's heads to regress the crowd count. Although the regression-based methods can regress the crowd counts directly, the location information of each person is omitted.

\begin{figure}
 
   \centering
   \includegraphics[width=1.0\linewidth]{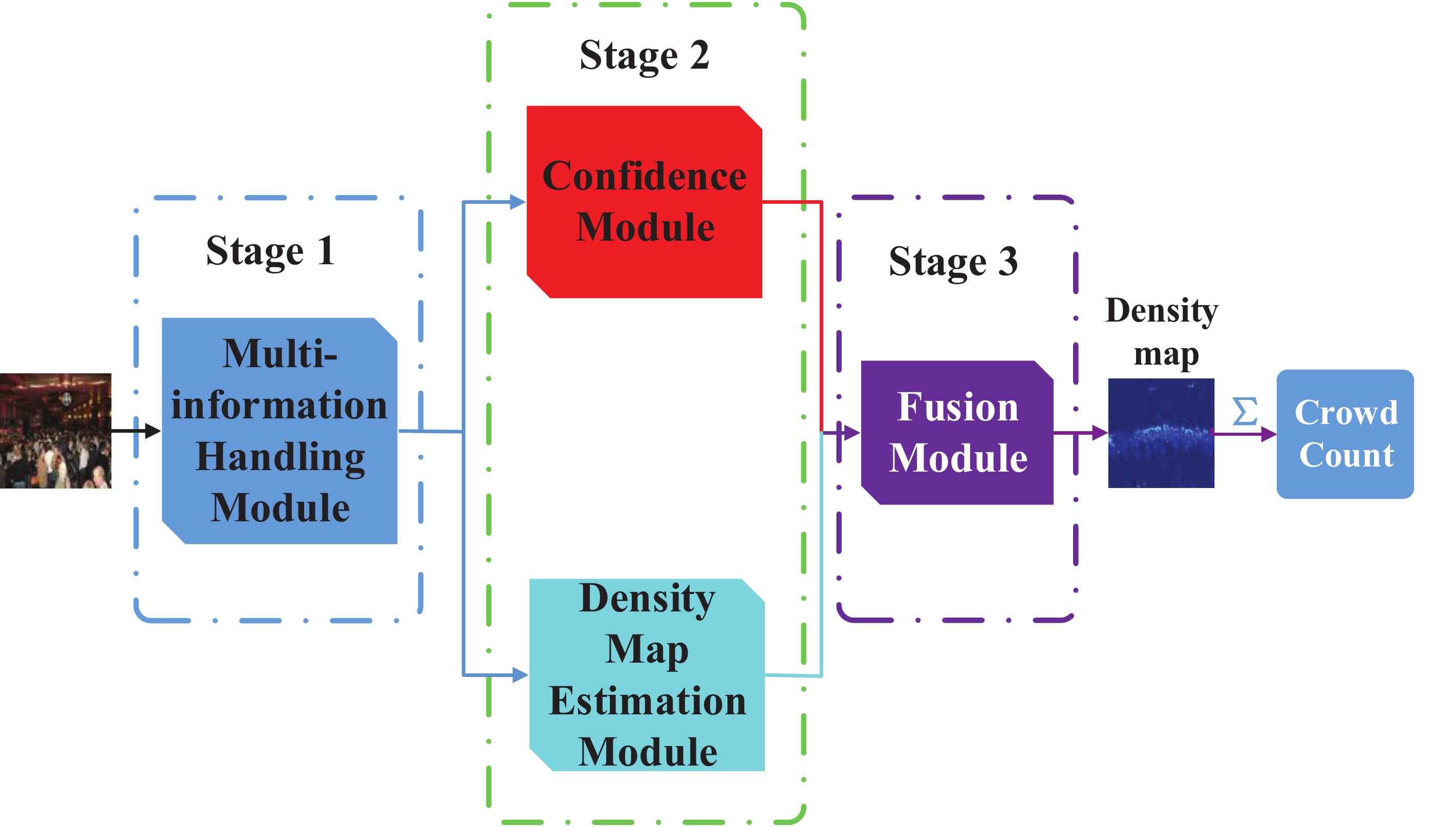}

   %\centering{\caption{The proposed architecture of our CAT-CNN}*}
  \caption{Overview of the proposed CAT-CNN. The Multi-information Handling Module is a feature extractor. The confidence map and estimated density map are generated respectively in the two middle modules. Then they are multiplied and further encoded to generate the high-precision density map in the Fusion Module.}
 
	\label{FIG:1}
\end{figure}

\begin{figure*}
   \centering
   \includegraphics[width=1.0\linewidth]{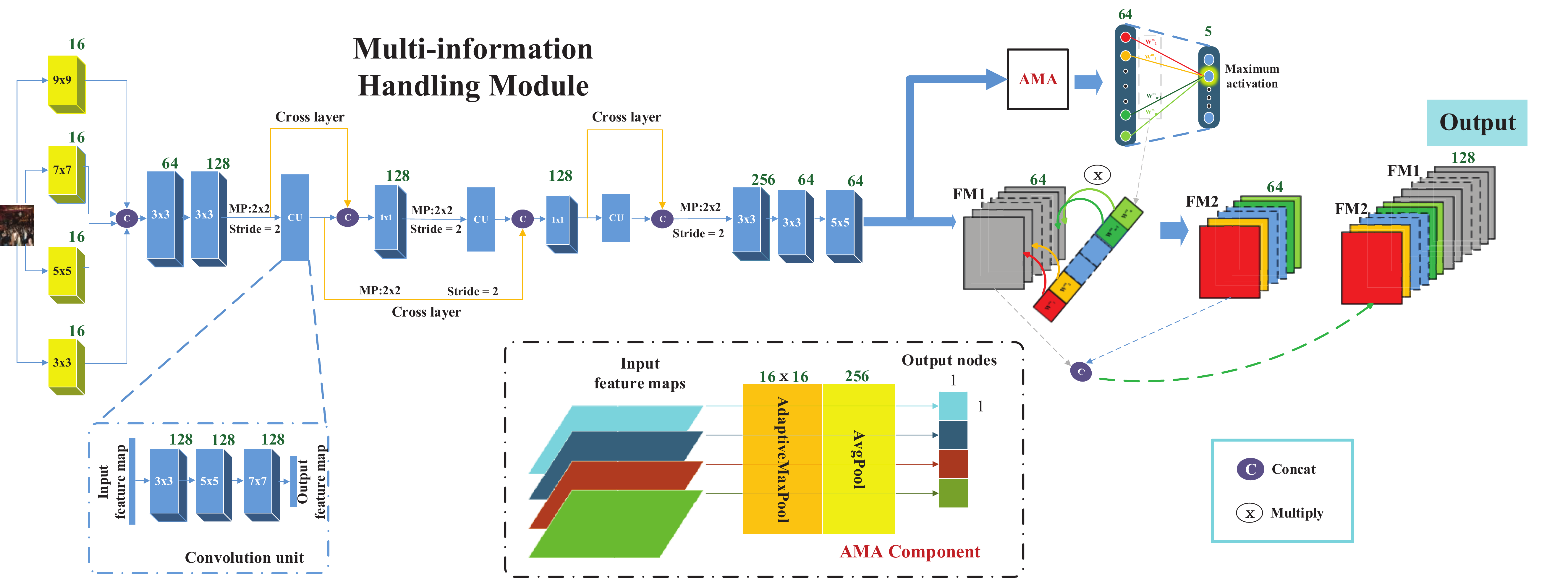}
   \caption {  The proposed architecture of the Multi-information Handling Module.}
	%\label{FIG:1}
\end{figure*}

\subsection{CNN-based methods}
Recently, due to the success of CNNs in many fields \cite{szegedy2015going,he2016deep,long2015fully,ren2015faster,hu2016dense}, CNN-based methods are widely used in crowd counting. The density map generated by CNNs records the count and location information of the crowd. Zhang et al. \cite{zhang2016single} proposed the MCNN to overcome scale variations. The MCNN leveraged three-branch CNNs with different convolution kernels to extract multi-scale features. Based on \cite{zhang2016single}, Sam et al. \cite{sam2017switching} designed the Switch-CNN with a classifier to select the optimal branch to encode a density map according to the variation of crowd counts. To employ the temporal correlation across frames in video sequences to assist crowd counting, Xiong et al. \cite{xiong2017spatiotemporal} introduced the ConvLSTM that could extract bidirectional timing information. Sindagi et al. \cite{sindagi2017generating} proposed the CP-CNN to incorporate global and local contextual features to encode a high-quality density map. Li et al. \cite{liu2018decidenet} proposed the DecideNet model to adaptively leverage the estimations of detection and regression. Zhang et al. \cite{zhang2019multi} proposed the MRA-CNN that could automatically focus on head regions by score maps. Hossain et al. \cite{hossain2019crowd} introduced a scale-aware attention mechanism to adapt the scale variation of crowds. Wang et al. \cite{Wang_2019_CVPR} designed the data collector and labeler to automatically generate and annotate the crowd data to reduce over-fitting caused by limited training data. In this paper, we propose a novel method to avoid misjudgements that can result in enormous errors of crowd counting.

% However, we find that due to the complexity of crowd scenes, these methods based on deep learning usually regard some objects as people by mistake, resulting in enormous errors of crowd counting. So we propose a novel method to solve the problem.

\section{Proposed methods}
\label{sec:method}

 %Our proposed CAT-CNN is shown in Fig. 2. The CAT-CNN contains four modules and three stages. Next, we will introduce these modules in detail. The Multi-information Handling Module is a novel architecture which aims to extract features from the input image to adapt to different scales and different crowd count groups, and the features are sent to Confidence Module and Density Map Estimation Module at the same time. The Confidence Module generates the confidence map to reflect the possibility about whether there are people at each pixel. The Density Map Estimation Module generates the estimated density map. The Fusion Module uses the confidence map to guide the estimated density map to generate the final density map. Next, we will introduce these modules in detail.

\subsection{Network architecture}
%\subsection{Multi-information Handling Module}
 An overview of the proposed CAT-CNN is shown in Fig. 2. Our CAT-CNN is composed of three stages. The first stage contains the first module where the features which can automatically adapt different scales and different crowd count groups are extracted. The second stage consists of two modules in the middle to encode confidence map and estimated density map respectively. The third stage contains the final module. With the guidance of the confidence map, final density map is encoded from the estimated density map in this stage. Next, we will elaborate these modules in each stage.

 %We use the design to help extract multi-scale features of the input image.
% In this paper, all of 9$\times$9 convolution represent the 3$\times$3 convolution with 4 dilation; all of 7$\times$7 convolution represent the 3$\times$3 convolution with 3 dilation; all of 5x5 convolution represent the 3$\times$3 convolution with 2 dilation. Every convolution is followed by rectified linear unit (ReLU) \cite{glorot2011deep}.

\noindent{\bfseries Multi-information Handling Module:} This module is proposed to overcome the scale variation and explicitly map the prior information of the population-level category back to feature maps to automatically contribute in encoding a highly refined density map. This module is illustrated in Fig. 3. To extract multi-scale features to overcome the scale variation, inspired by MCNN \cite{zhang2016single}, we exploit four different kernels to convolve the input image at the beginning of this module. Besides, several 2$\times$2 max-pooling layers are designed inside this module. And we fuse convolutional layers of different depths to fully excavate the multi-scale features. To alleviate overfitting caused by redundant parameters, we widely employ the dilated convolution \cite{yu2015multi} in our CAT-CNN which can expand the receptive fields of convolution with fewer parameters. In this paper, all convolutions of different shapes are constituted by the 3$\times$3 convolution with corresponding dilation, except for the 1$\times$1 convolution. Every convolution is followed by a rectified linear unit (ReLU) \cite{glorot2011deep}.

%In the crowd count group classifier, the total crowd count in each image is classified \cite{li2018pdisvpl}.
   Inspired by \cite{sindagi2017cnn}, the crowd counts are quantized into five groups in each dataset and a crowd count group classifier is learned. A simple example about the process of statistics and classification of datasets is shown in Fig. 5. In the crowd count group classifier, the total crowd count of each image is classified. To feed arbitrarily sized images into the fully-connected (FC) layer for training without resizing images to maintain the original distribution of the crowd, inspired by \cite{lin2013network,sindagi2017cnn}, we employ AdaptiveMaxPool and AvgPool (AMA) to design a novel model named AMA Component. As shown in Fig. 3, arbitrarily sized input can be fed to the AMA Component. And the output is always 1$\times$1 node. This component is placed between convolutional layer and FC layer to form the novel classification model. Inspired by \cite{zhou2016learning}, the major improvement is that AMA has one more AvgPool layer than SPP used in \cite{sindagi2017cnn,yang2018counting}. The purpose of this design is to directly map the weights of the predicted population-level category back to feature maps, which is also the main distinction from \cite{sindagi2017cnn,yang2018counting}. We first explicitly map the prior information of the population-level category back to feature maps to automatically contribute in encoding a highly refined density map.

\begin{figure*}

   \centering
   \includegraphics[width=1.0\linewidth]{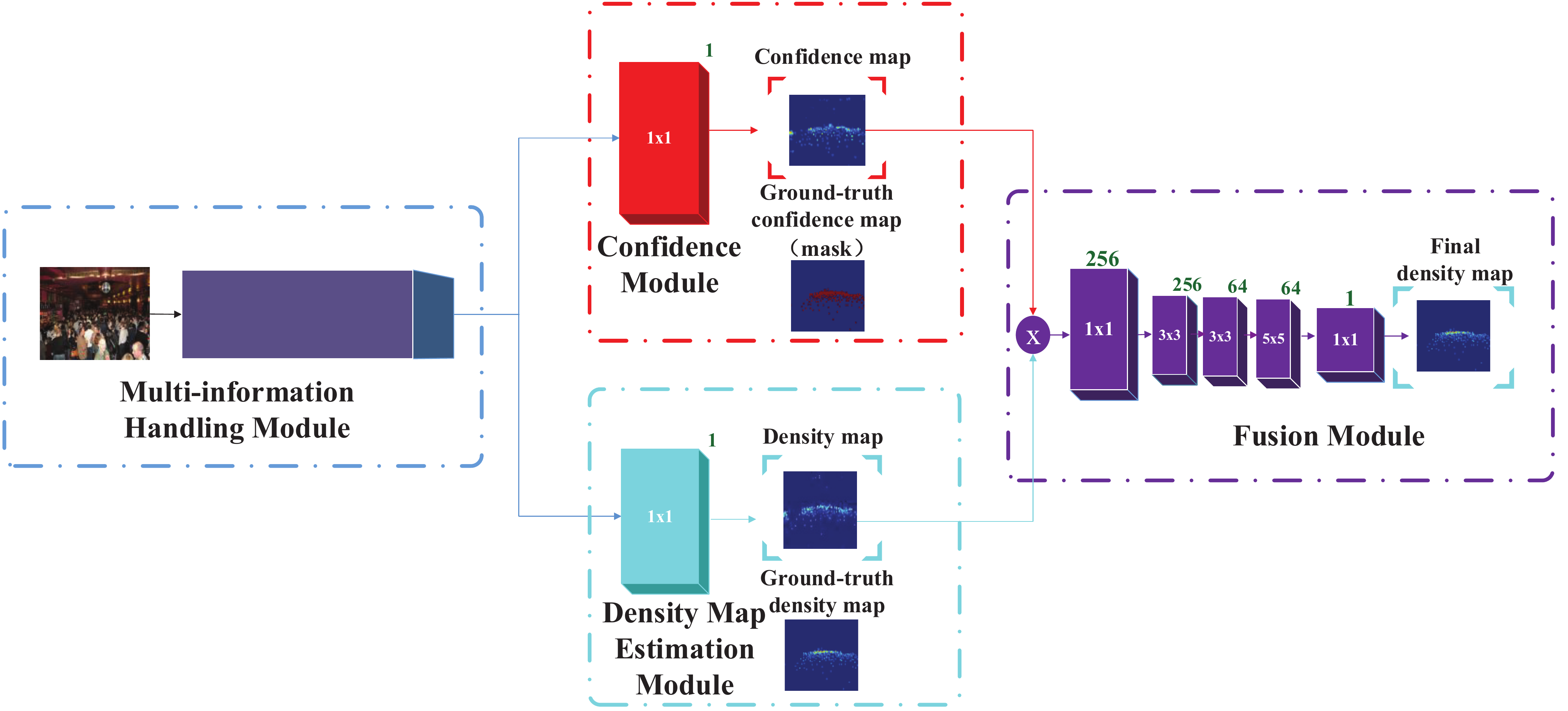}

   %\centering{\caption{The proposed architecture of our CAT-CNN}*}
  \caption{The proposed architecture of our CAT-CNN.}
 
	\label{FIG:1}
\end{figure*}

\begin{figure}

   \centering
   \includegraphics[width=0.8\linewidth]{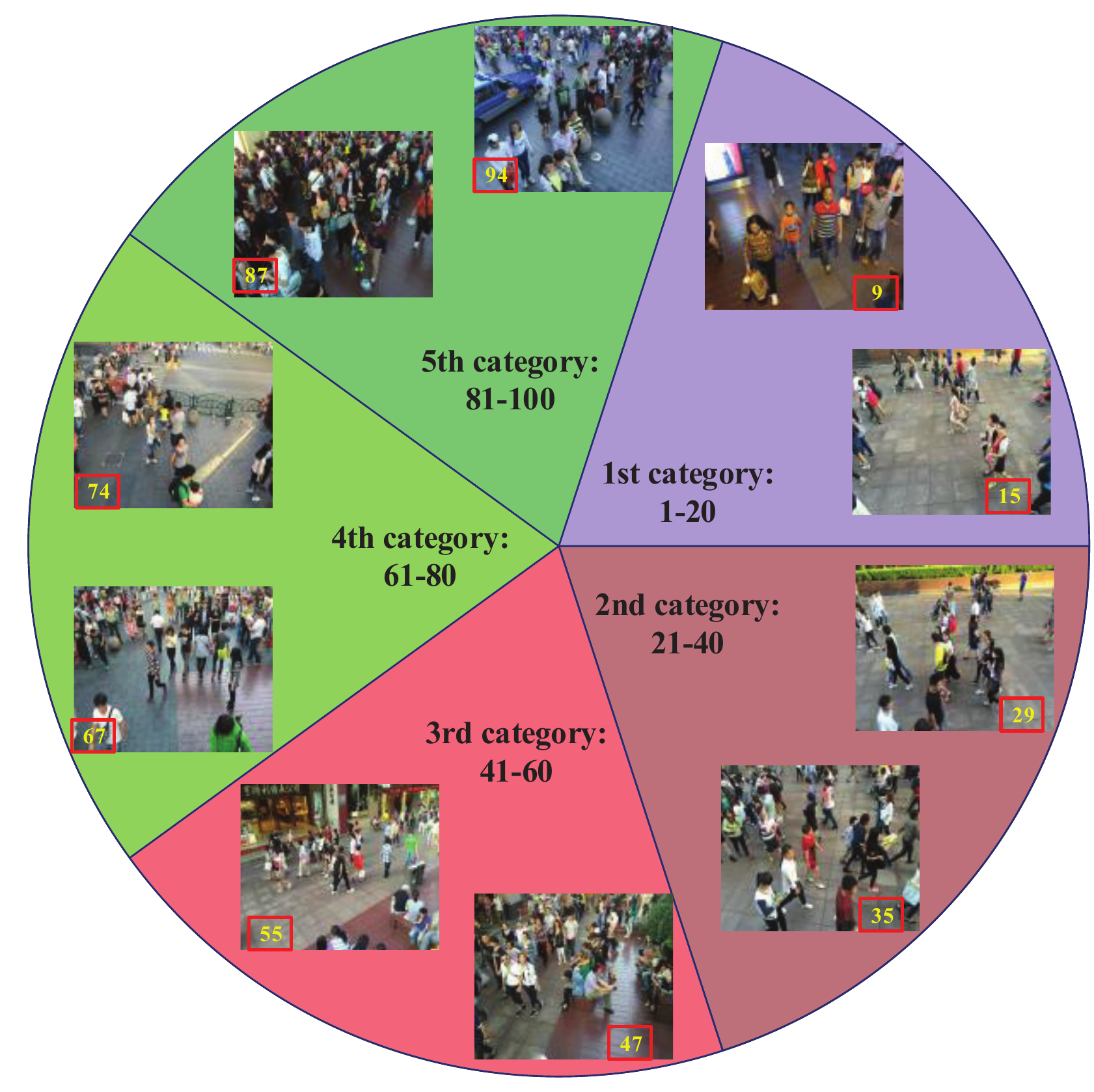}

   %\centering{\caption{The proposed architecture of our CAT-CNN}*}
   \centering{\caption{A simple example about the process of statistics and classification of datasets: The crowd counts range from 1 to 100 in a dataset and they are quantized into five groups. The crowd count group classifier can divide the image into corresponding groups according to the crowd count in the image. The number in red box represents the crowd count in images.}}
 
	\label{FIG:1}
\end{figure}

 As shown in Fig. 3, we leverage the feature map (FM1) to feed the AMA Component and outputs are sent to the FC layer followed by Parametric ReLU \cite{he2015delving}. The maximum activated neuron in FC layer represents the population-level category of input image. We explicitly use the prior of the population-level category by multiplying the weights of predicted category to feature maps (FM1). The mapped feature maps (FM2) can be used as a feature vector representation to characterize the population-level category. FM2 is concatenated with FM1 to serve as the output of this module to retain more sufficient feature information for following modules. We choose minimizing the cross-entropy loss to optimize the novel classification task.

     %is helpful to reduce the crowd counting errors. At the end of the Multi-information Handling Module, the FC layer is connected, which can classify the crowd into five count groups to adapt to different crowd count distributions. 

   %In order to avoid the distortion of images and maintain original distribution information of the crowd, we do not resize the training images. So the Spatial Pyramid Pooling (SPP) \cite{he2015spatial} layer is placed between convolutional layers and fully connected layers. The SPP is proposed by the team of Kaiming He. Arbitrarily sized convolutional features can be fed to the SPP layer, and the SPP layer can produce fixed size outputs to feed the FC layers. We choose minimizing the cross-entropy error to optimize this process.

%\subsection{Confidence Module}

\noindent{\bfseries Confidence Module:} Due to the complexity of crowd scenes, CNN-based methods often mistake some objects as human heads. If the prior of whether there are people at each pixel location can be used, this problem will be solved easily. We propose the Confidence Module based on probability. The probability of each pixel belonging to a person's head can be estimated by dense classifications.
%Our motivation is simple but useful: the deep learning network usually regards some objects as people's head and count mistakely, the main reason is that it forms wrong judgements about whether there are people at this location. Hence, using classification to judge whether there are people at each pixel location can solve the problem effectively.  

%solve the problem of unbalanced population distribution, we propose weighted BCEloss to obtain a robust confidence map for population distribution.

   The proposed Confidence Module is shown in Fig. 4. The output of the Multi-information Handling Module is fed into this module. With the supervision of ground-truth confidence map, the 1$\times$1 convolution followed by ReLU is utilized to encode the feature maps and transform them into predicted confidence map. The ground-truth confidence map is a binary map. We set the probability of the pixel belonging to a human head to 1. The predicted confidence map records the estimated possibility of a human head at each pixel location. The greater probability in the predicted confidence map, the more attention our model will pay in the density map by multiplying the confidence map in the Fusion Module. Due to the unbalanced population distribution, we propose the weighted BCEloss to obtain a robust confidence map for population distribution. The weighted BCEloss will be elaborated in Sec. 3.2.
%The weighted BCEloss to reduce the difference between the confidence map and its ground truth. In order to distinguish its ground truth from other modules’ ground truth, we call this ground truth as ground-truth confidence map (mask). It will be introduced in Sec. 4.1.

%\subsection{Density Map Estimation Module}

%In crowd counting, the people are seriously obscured on high density scenes. It is difficult to get the precise number of people by detecting the whole body. So we complete the challenging task by counting the head of people which can significantly figure out the difficult problems. The Density Map Estimation Module can get the estimated density map of people's heads.

% In the crowd counting task, some people are seriously obscured in the high density scene. It is difficult to get the precise number of people by detecting the whole body. We address this problem by counting the head of people.

\noindent{\bfseries Density Map Estimation Module:} The estimated density map of human heads is encoded in this module. As shown in Fig. 4, the output of the Multi-information Handling Module is fed into this module. With the supervision of ground-truth density map, the 1$\times$1 convolution followed by ReLU is utilized to encode the feature map to the estimated density map. The ground-truth density map will be elaborated in Sec. 3.2. Euclidean loss is used to optimize the estimated density map.

%\subsection{Fusion Module} 

\noindent{\bfseries Fusion Module:} Firstly, we distinguish the difference between confidence map and estimated density map in detail. The classification-based confidence map reflects the possibility of a human head at each pixel location. But it can't be used to count the crowd directly. The estimated density map contains the pivotal information of crowd counts. However, it usually regards some objects as human heads mistakenly. We combine both to avoid the misjudgements in crowd counting. As shown in Fig. 4, the estimated density map is multiplied by confidence map and the results are sent to the following convolutional layer and ReLU. With the guidance of the confidence map, the position of human head in the estimated density map gets more attention to encode the final density map. Then enormous misjudgements can be avoided in the final density map. Euclidean loss is used to optimize the final density map. The crowd count can be obtained by integrating the final density map. In \cite{zhang2019multi}, its attention mechanism is also important in MRA-CNN, which inspires us to explore the attention mechanism of MRA-CNN in the future.

%As shown in Fig. 6, firstly, the Fusion Module uses 1x1 kernel size to upgrade the channel number of confidence map and estimated density map. Then they are multiplied and the result is sent to the following convolutional layers. At the end of model, we use 1x1 kernel size to produce the final density map, the Euclidean distance loss is used to optimize the final density map, and integrating the final density map can get the crowd count. The final density map’s ground truth is the same as the ground truth of the Density Map Estimation Module.  It can be used to avoid enormous wrong judgements.

\subsection{Implementation}
\label{ssec:train}

%\subsection{Ground truth generation}
\noindent{\bfseries Ground truth generation:} Following the method in \cite{sindagi2017cnn}, the head positions $P$ are provided in each image $I_{i}$. All of annotated points $A_{i}$ are convolved by a Gaussian kernel centered on each annotated point to encode the ground-truth density map. The Gaussian kernel with $\sigma$ = 4.0 has been normalized to 1. The density of a specific pixel $p$ in one image can be regarded as the Gaussian function effects of its surrounding effective annotated points:\begin{center}
\begin{equation}
 p\in I_{i} , D(p)=\sum _{P\in A_{i}}N(p;\mu _{i},\sigma ^{2}I_{2\times2}),     
\end{equation}
\end{center}where $N$ represents the response of a 2D Gaussian function with its mean $\mu _{i}$. $I_{2\times2}$ represents its isotropic $2\times2$ covariance matrix with variance $\sigma ^{2}$. We can integrate the density value at each pixel over the entire map to get the crowd count $\sum _{p\in I_{i}} D(p) = Num_{i} $. %The crowd counting module can learn a non-linear mapping for the training images. We use above simple method to create the ground-truth density map in order to ensure that the improvement of result comes from our novel method instead of the innovation of creating ground truth of density maps.

\begin{figure}
	\centering
   \includegraphics[width=1.0\linewidth]{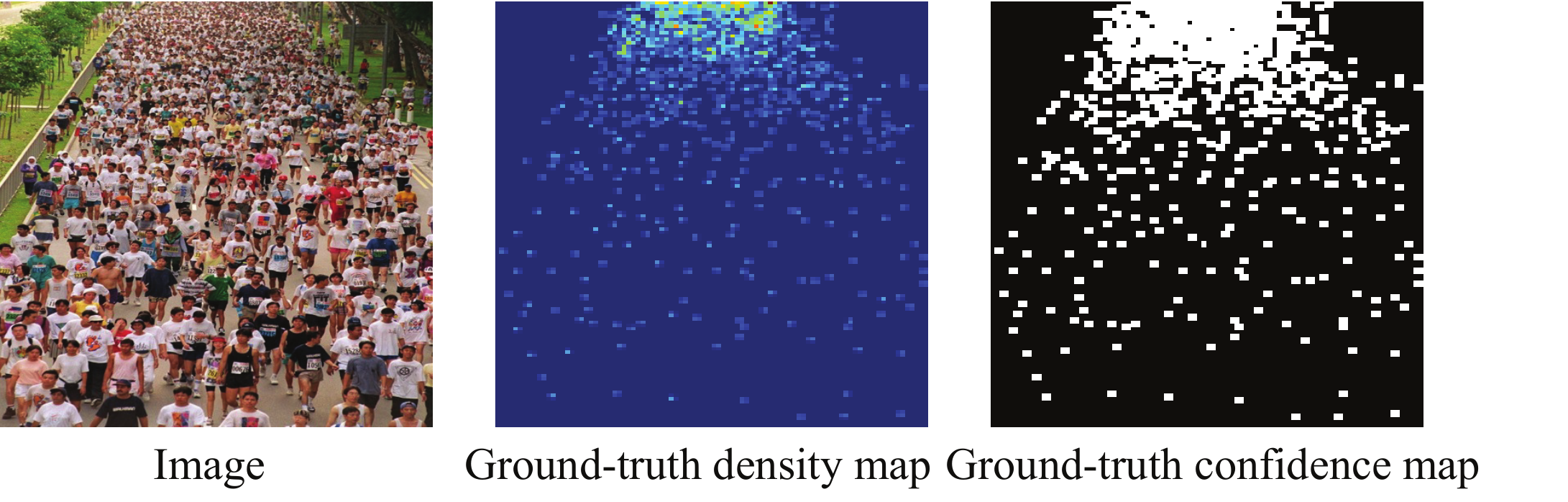}

   \caption{ Visualization results of the ground-truth density map and ground-truth confidence map (mask).}
	\label{FIG:1}
\end{figure}

%with a shape size of 15$\times$15
Due to the enormous number of people in datasets, the cost of manually labeling the ground-truth confidence map is expensive. For example, the ShanghaiTech Part\_A \cite{zhang2016single} dataset contains 241,677 people. We propose a simple but effective method to obtain the ground-truth confidence map. Specifically, the ones template where all pixels are set to 1 is first generated. Then the ones template centered on each annotated point is pasted on a binary map to encode the ground-truth confidence map (mask). The ones template and the Gaussian kernel have the same size that is set to 15$\times$15 in \cite{sindagi2017cnn}. The ground-truth confidence map (mask) generation uses the same size on different datasets. In the ground-truth confidence map (mask), '1' represents 100$\%$ possibilities of a human head at this pixel location and '0' represents 0 possibility of a human head at this pixel location. The BCEloss that can predict the probability of a positive sample is utilized to optimize the confidence map by supervised learning. The greater probability in the predicted confidence map, the more attention our model will pay in the density map by multiplying the confidence map. Visualization results are shown in Fig. 6.

%\subsection{Module Optimization} 
\noindent{\bfseries Module Optimization:} Our CAT-CNN contains four modules. The whole loss function $L_{whole}$ is given by Eq.(2):\begin{center}
\begin{equation}
 L_{whole} =L_{fus} +  L_{den} +  \lambda_{1}L_{con} + \lambda_{2}L_{mul},
\end{equation}
\end{center}where $L_{mul} $,  $L_{con}$,  $L_{den} $, and  $L_{fus} $ are the losses for Multi-information Handling Module, Confidence Module, Density Map Estimation Module, and Fusion Module, respectively. $\lambda_{1}$ and $\lambda_{2}$ are the hyper-parameters to balance the magnitude of two tasks. $\lambda_{1}$ is set to 2 and $\lambda_{2}$ is set to 1e-2.

   In the Multi-information Handling Module, we classify the crowd into five groups according to the crowd counts in each dataset. The cross-entropy loss function is used in this module:
\begin{center}
\begin{equation}
L_{mul} = -\frac{1}{N}\sum ^{N}_{i=1}\sum ^{K}_{j=1}\left [ \left ( q^{i}=j \right )G_{mul}\left ( X_{i},\theta \right ) \right ],
\end{equation}
\end{center}
where $N$ represents the total number of training images.  $\theta $ represents a set of network parameters. $X_{i} $ represents the $i^{th} $ training image. $q^{i} $ represents the ground-truth class and $K$ represents the whole number classes. $G_{mul}\left ( X_{i},\theta \right ) $ represents the output of classification.

   In the Confidence Module, we first propose employing classification at each pixel location to judge whether there are people. Due to the unbalanced population distribution in images, we propose the weighted BCELoss to encode a robust confidence map for population distribution. The weighted binary-cross entropy loss adopted in this module is shown as follows:

\begin{center}
\begin{equation}
 %\! L_{att}\!  = \! - w_{i}\left[ y_{i}\log \left( h_{\theta}\left( X_{i} \right ) \right )\! +\! \left( 1 - y_{i} \right )\log \left( 1 - h_{\theta}\left( X_{i} \right )\! \right )\right ], \!
  L_{con}  = - w_{i}\left[ y_{i}\log  X_{i}  + \left( 1 - y_{i} \right )\log \left( 1 -  X_{i} \right )\right ], 
\end{equation}
\end{center}where $X_{i}$ represents the confidence map that records the predicted probability of a positive sample. $y_{i}$ indicates the ground-truth confidence map (mask). $w_{i}$ is the weight given to the loss of each element. The derivation process of $w_{i}$ is given as follows:
\begin{center}
\begin{equation}
\begin{aligned}
\quad \quad \quad \quad \quad w_{i} &= f_{i}\cdot \bar{y}_{i} + b_{i}\cdot y_{i},\\
\quad \quad \quad \quad \quad f_{i} &= mean(y_{i}),\\
\quad \quad \quad \quad \quad b_{i} &= 1 - f_{i},\\
\quad \quad \quad \quad \quad \bar{y}_{i} &= 1 - y_{i}.
\end{aligned}
\end{equation}
\end{center}

\begin{center}
\begin{equation}
\begin{aligned}
w_{i} = \left\{\begin{matrix} &10^{-6}, &if \,\,\, f_{i}=0; \\ &(1- 2f_{i})y_{i}+f_{i}, &else. \end{matrix}\right.
\end{aligned}
\end{equation}
\end{center}

Where $f_{i}$ represents the weight of crowd positions, and $b_{i}$ represents the weight of background positions. $\bar{y}_{i}$ represents the ground truth of  background. For better convergence of our CAT-CNN, when $f_{i}$ = 0 which means that there is no person in the training image, we set $w_{i}$ to $10^{-6}$ instead of 0.

\begin{table*}
\caption{Statistics of training datasets: Num represents the number of images; Range represents the range of crowd counts; Average represents the average crowd counts; Total represents the total number of labeled people.}

\begin{center}
\label{my-label}
\begin{tabular}{cccccccc|}\hline
Dataset                                & Resolution & Color & Num & Range & Average & Total \\ \hline 

UCF\_CC\_50               & different   & Grey  & 50   & [94,  4543]  & 1279.5   & 63,974 \\ 
ShanghaiTech Part\_A           & different   & RGB,Grey  & 482  & [33, 3139]  & 501.4   & 241,677 \\ 
ShanghaiTech Part\_B                   & 768 x 1024   & RGB  & 716  & [9,  578]  & 123.6   & 88,488 \\
WorldExpo’10           & 576 x 720   & RGB  & 3980  & [1,  253]  & 50.2   & 199,923 \\ \hline
\end{tabular}
\end{center}
\end{table*} 

   In the Density Map Estimation Module and Fusion Module, the Euclidean distance loss is used to optimize these modules:
\begin{center}
\begin{equation}
\begin{aligned}
L_{den} = \frac{1}{2M}\sum _{i=1}^{M}\left \| D\left ( X_{i} ,\theta \right ) - D_{i} \right \|^{2}_{2},\\
L_{fus} = \frac{1}{2M}\sum _{i=1}^{M}\left \| F\left ( X_{i} ,\theta \right ) - D_{i} \right \|^{2}_{2}, 
\end{aligned}
\end{equation}
\end{center}
where $D\left ( X_{i} ,\theta \right ) $ represents the estimated density map in the Density Map Estimation Module. $ F\left ( X_{i} ,\theta \right ) $ represents the final density map in the Fusion Module. These two modules have the same ground-truth density map $ D_{i} $. $M$ represents the total number of training samples.

Each module has its subtask. These four subtasks are completed synchronously in a synergistic manner. The optimized process is a multi-task learning \cite{collobert2008unified,wang2016large} which is helpful to reduce overfitting caused by limited data for training. Multi-task learning can extract the generalizable representation of similar data and assist our model in alleviating crowd counting errors. \cite{yang2018counting} also adopts the strategy of multi-task learning. There are two main differences between two works. Firstly, our proposed CAT-CNN is a multi-stage model. The confidence map is encoded in the second stage. In the third stage, the predicted confidence map is multiplied by the estimated density map to directly participate in the generation of final density map to avoid enormous misjudgements. In \cite{yang2018counting}, the BG/FG mask is not directly involved in density map generation, but fine-tunes the density map as a multi-task learning. The rationality of two designs is guaranteed by different loss functions. Secondly, for the classification task in Fig. 3, inspired by \cite{zhou2016learning}, we design the AMA Component and the weights of the predicted category are directly mapped back to previous feature maps to contribute in generating a highly refined density map in our work, which is the second difference from \cite{yang2018counting}.

%\begin{center}
%\begin{equation}
%L_{fus} = \frac{1}{2M}\sum _{i=1}^{M}\left \| F\left ( X_{i} ,\theta \right ) - D_{i} \right \|^{2}_{2}, 
%\end{equation}
%\end{center}
%where $ F\left ( X_{i} ,\theta \right ) $ represents the final estimated density map in Fusion Module.

\begin{figure}
\centering
   \includegraphics[width=1.0\linewidth]{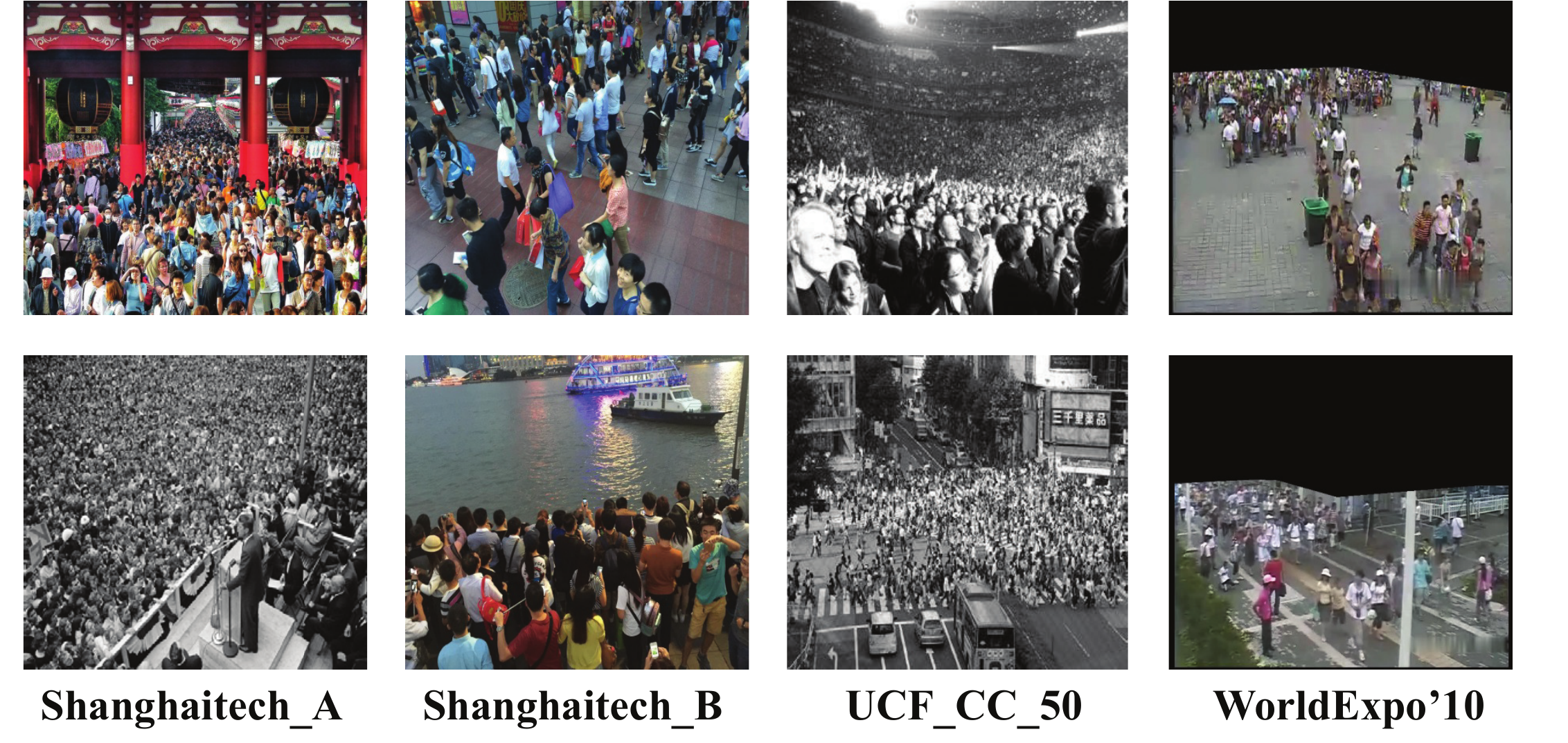}
   \caption{Some representative examples from these datasets.}
	%\label{FIG:1}
\end{figure}

\section{Experiments}
\label{sec:experi}

%\subsection{Experiment setting} 
%\subsection{Evaluation Metric} 
Our model is trained on three benchmark datasets separately: Shanghaitech dataset \cite{zhang2016single}, UCF\_CC\_50 dataset \cite{idrees2013multi}, and WorldExpo'10 dataset \cite{zhang2015cross}. Statistics of them are summarized in Table 1. Some example images of these datasets are shown in Fig. 7. The data augmentation schemes in \cite{sindagi2017cnn} are utilized to reduce overfitting. In the process of creating training datasets, 9 patches are cropped from each original image at random locations, and the size of each patch is $1/4^{th}$ of the original image. Random flipping and uniform noise are used in the patch with a probability of 0.5.  The Adam optimization algorithm with a batch size of 1 is used to optimize these loss functions in our model. Its parameter of learning rate is set to 1e-6. The number of epochs is set to 5000. In every iteration, gradients for each loss function are calculated and corresponding parameters are updated. The curve of each loss function is expected to converge to smaller values. It may have a brief shock at the beginning. The experiments are conducted on NVIDIA GTX 1080Ti and Intel Core i7 with the Torch framework.

  Following existing methods \cite{zhang2016single,sam2017switching,sindagi2017generating,sam2018top,liu2018decidenet,shen2018crowd,shi2018crowd,babu2018divide}, the Mean Absolute Error (MAE) and Mean Squared Error (MSE) are used to measure the count errors. They are defined as:
\begin{center}
\begin{equation}
\! MAE \!= \!\frac{1}{N}\sum _{i=1}^{N}\left | z_{i}-\widetilde{z_{i}} \right |, MSE\! = \!\sqrt{\frac{1}{N}\sum ^{N}_{i=1}\left ( z_{i} - \widetilde{z_{i}} \right )^{2}}. 
\end{equation}
\end{center}  

%\begin{center}
%\begin{equation}
% MSE = \sqrt{\frac{1}{N}\sum ^{N}_{i=1}\left ( z_{i} - \widetilde{z_{i}} \right )^{2}}. 
%\end{equation}
%\end{center}
Where $N$ is the number of test images. $z_{i}$ is the actual crowd count by integrating the ground-truth density map, and $\widetilde{z_{i}}$ is the estimated crowd count by integrating the final density map.

\begin{table*}
\setlength{\abovecaptionskip}{0.4cm}
\setlength{\belowcaptionskip}{-0.1cm}
\begin{center}
\centering
%\caption{Estimation errors on ucf_cc_50 dataset}
\caption {Estimation errors (MAE) of different configurations about the number of population-level categories.} 

\label{my-label}
\begin{tabular}{cccccc}
\hline
                               & \multicolumn{5}{c}{population-level categories} \\  \cmidrule(lr){2-6}
                         & 3      & 5     & 7      & 10     & 15                 \\ \hline

ShanghaiTech Part\_A       & 78.2    & \textbf{66.7}  & 69.8  & 71.8    & 71.3         \\
ShanghaiTech Part\_B       & 15.3    & \textbf{11.2}  & 13.2  & 12.9    & 12.2         \\
UCF\_CC\_50                & 308.4    & \textbf{235.5}  & 280.3  & 301.6    & 340.4         \\
WorldExpo’10              & 8.6   & 7.2  & 8.0  & 8.5    & \textbf{7.1}       \\ \hline
\end{tabular}
\end{center}
%\caption{Estimation errors for different configurations on ShanghaiTech Part\_A.}

\end{table*}

\begin{table*}
\setlength{\abovecaptionskip}{0.4cm}
\setlength{\belowcaptionskip}{-0.1cm}
\begin{center}
\centering
%\caption{Estimation errors on ucf_cc_50 dataset}
\caption{Estimation errors of different kernel sizes at the beginning of CAT-CNN.}

\label{my-label}
\begin{tabular}{ccccccccc}
\hline
                                                             & \multicolumn{2}{c}{Part\_A}    & \multicolumn{2}{c}{Part\_B}    & \multicolumn{2}{c}{UCF\_CC\_50}   & \multicolumn{2}{c}{WorldExpo’10}  \\  \cmidrule(lr){2-3} \cmidrule(lr){4-5}  \cmidrule(lr){6-7} \cmidrule(lr){8-9}
                                                             & MAE           & MSE              & MAE           & MSE                 & MAE           & MSE               & MAE           & MSE         \\ \hline
3$\times$3                                                   & 76.8        & 125.9                & 14.3        & 24.1                 & 289.6        & 441.5               & 8.4        & 11.3         \\ 
3$\times$3,5$\times$5,7$\times$7                             & 78.2         & 128.1               & 13.4        & 22.7                & 306.4        & 465.6               & 9.2       & 13.8 \\ 
3$\times$3,5$\times$5,7$\times$7,9$\times$9                  & \textbf{66.7}& \textbf{101.7}      &\textbf{11.2} & \textbf{20.0}      &\textbf{ 235.5}  & \textbf{324.8}    & \textbf{7.2}       & 9.5 \\ 
3$\times$3,5$\times$5,7$\times$7,9$\times$9,11$\times$11     & 73.8         & 120.3                & 12.8        & 22.8                & 271.8        & 370.2               & 7.4  & \textbf{9.3}\\ \hline

\end{tabular}
\end{center}
%\caption{Estimation errors for different configurations on ShanghaiTech Part\_A.}

\end{table*}

\begin{table*}
\setlength{\abovecaptionskip}{0.4cm}
\setlength{\belowcaptionskip}{-0.1cm}
\begin{center}
\centering
%\caption{Estimation errors on ucf_cc_50 dataset}
\caption{Estimation errors of different configurations about the cross-layer connection.}

\label{my-label}
\begin{tabular}{ccccccccc}
\hline
                                                             & \multicolumn{2}{c}{Part\_A}    & \multicolumn{2}{c}{Part\_B}    & \multicolumn{2}{c}{UCF\_CC\_50}   & \multicolumn{2}{c}{WorldExpo’10}  \\  \cmidrule(lr){2-3} \cmidrule(lr){4-5}  \cmidrule(lr){6-7} \cmidrule(lr){8-9}
                                                             & MAE           & MSE              & MAE           & MSE                 & MAE           & MSE               & MAE           & MSE         \\ \hline
Without cross-layer connection                               & 75.5        & 124.9                & 16.3        & 28.2                 & 264.2       & 382.8               & 10.1        & 14.2         \\ 
With cross-layer connection                & \textbf{66.7}         & \textbf{101.7}           & \textbf{11.2}       & \textbf{20.0} & \textbf{235.5}       & \textbf{324.8}     & \textbf{7.2}        & \textbf{9.5} \\ \hline

\end{tabular}
\end{center}
%\caption{Estimation errors for different configurations on ShanghaiTech Part\_A.}

\end{table*}

\begin{table}
\setlength{\abovecaptionskip}{0.4cm}
\setlength{\belowcaptionskip}{-0.1cm}
\begin{center}
\centering
%\caption{Estimation errors on ucf_cc_50 dataset}
\caption {Estimation errors of different configurations about the output of Multi-information Handling Module on ShanghaiTech Part\_A. } 

\label{my-label}
\begin{tabular}{cccccc}
\hline
                            
Method                       & MAE           & MSE                     \\ \hline

Only FM1  & 76.7         & 115.3         \\ 
Only FM2  & 71.3         & 109.1         \\ 
FM1 and FM2   &\textbf{66.7}         & \textbf{101.7}         \\ \hline    
\end{tabular}
\end{center}
%\caption{Estimation errors for different configurations on ShanghaiTech Part\_A.}

\end{table}

\begin{table}
\setlength{\abovecaptionskip}{0.4cm}
\setlength{\belowcaptionskip}{-0.1cm}
\begin{center}
\centering
%\caption{Estimation errors on ucf_cc_50 dataset}
\caption {Estimation errors of different configurations about the confidence map on ShanghaiTech Part\_A. } 

\label{my-label}
\begin{tabular}{cccccc}
\hline
                            
Method                       & MAE           & MSE                     \\ \hline

Without confidence map  & 81.2         & 128.9         \\ 
With confidence map      &\textbf{66.7}         & \textbf{101.7}         \\ \hline    
\end{tabular}
\end{center}
%\caption{Estimation errors for different configurations on ShanghaiTech Part\_A.}

\end{table}

\begin{table}
\setlength{\abovecaptionskip}{0.cm}
 
\setlength{\belowcaptionskip}{-0.9cm}

\caption{Comparison of real time in Cascade-MTL, MRA-CNN, and CAT-CNN. Time represents the time to process a frame.}

\begin{center}
\scalebox{0.80}{
\label{my-label}
\begin{tabular}{ccccccc}
\hline
Method                                                          & \multicolumn{6}{c}{Time (s)}  \\  \cline{2-7}
                                                               & S1 & S2 & S3 & S4 & S5 & Average \\ \hline
Cascade-MTL  \cite{sindagi2017cnn}                             & 0.08   & 0.08  & 0.08  & 0.08  & 0.08   & 0.08 \\    %\cite{sindagi2017cnn} 
MRA-CNN \cite{zhang2019multi}                                 & 0.02  & 0.02  & 0.02  & 0.02 & 0.02  & 0.02 \\  %\cite{zhang2019multi} 

CAT-CNN(OURS)                                   & 0.03  & 0.03 & 0.03  & 0.03  &0.03   & 0.03  \\ \hline
\end{tabular}}
\end{center}
%\caption{Estimation errors on the WorldExpo'10 dataset.}
\end{table}

\begin{figure*}
\setlength{\abovecaptionskip}{1.5cm}
\setlength{\belowcaptionskip}{-0.1cm}

   \centering
   \includegraphics[width=1.0\linewidth]{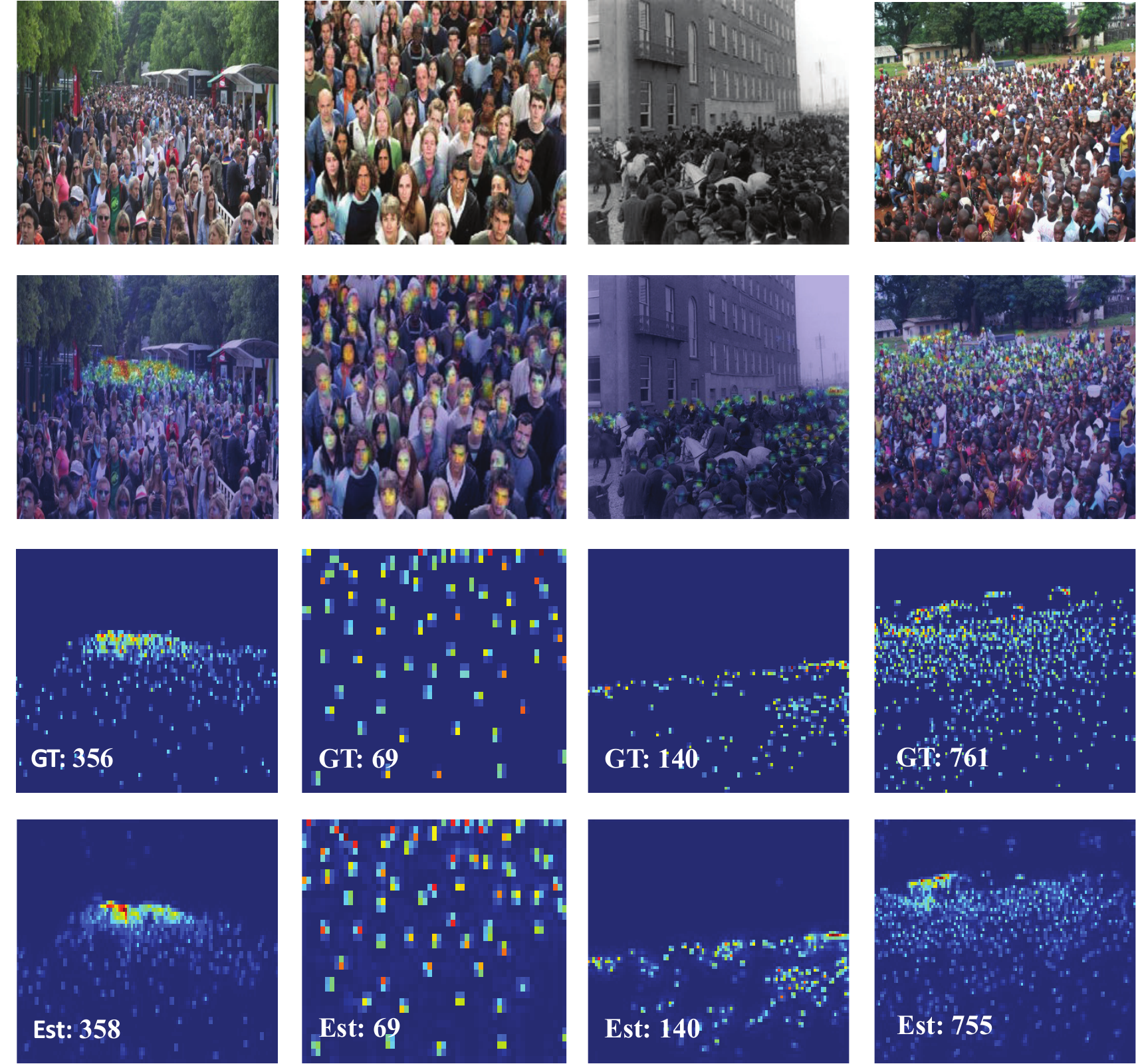}

    \centering
   %\centering{\caption{The proposed architecture of our CAT-CNN}*}
  \caption{Results of the proposed CAT-CNN on ShanghaiTech Part\_A. First Row: test image. Second Row: test image overlaid by the estimated confidence map. Third Row: ground-truth density map. Fourth Row: the final estimated density map.}
 
	\label{FIG:1}
\end{figure*}

\begin{figure*}

   \centering
   \includegraphics[width=1.0\linewidth]{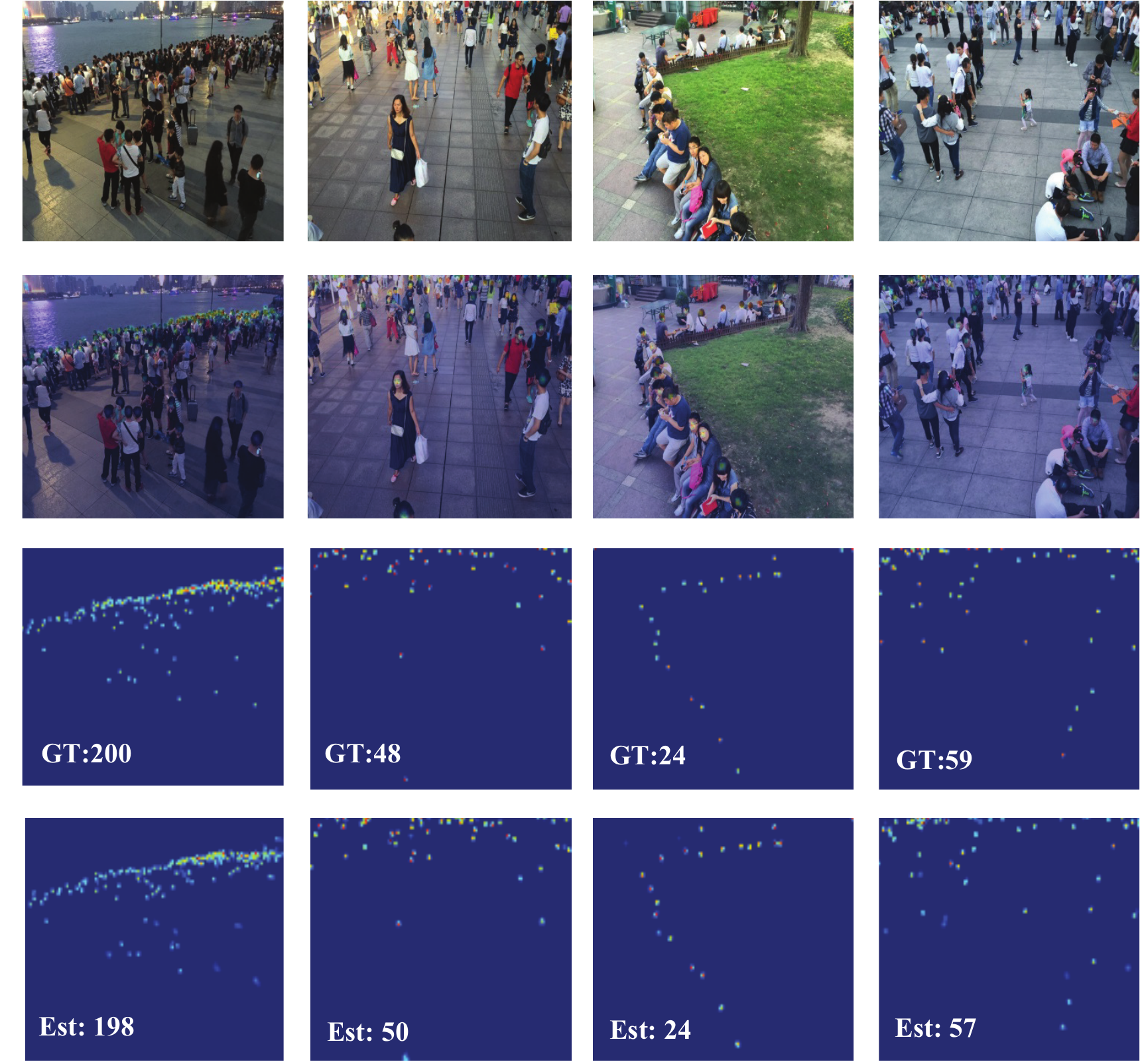}

   %\centering{\caption{The proposed architecture of our CAT-CNN}*}
   \centering{\caption{Results of the proposed CAT-CNN on ShanghaiTech Part\_B. First Row: test image. Second Row: test image overlaid by the estimated confidence map. Third Row: ground-truth density map. Fourth Row: the final estimated density map.}}
 
	\label{FIG:1}
\end{figure*}

\begin{figure}

   \centering
 \includegraphics[width=1.0\linewidth]{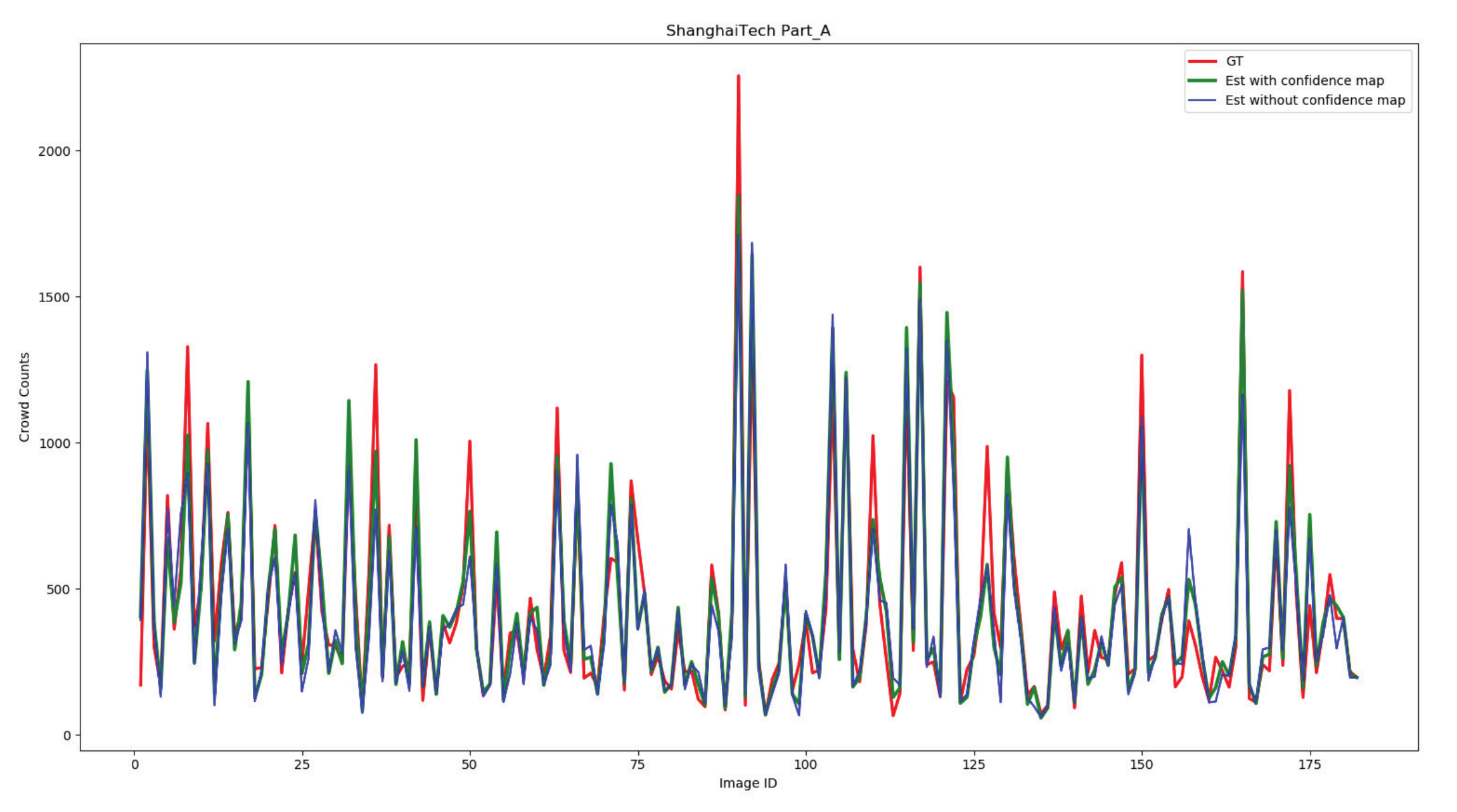}

   %\centering{\caption{The proposed architecture of our CAT-CNN}*}
   \centering{\caption{The crowd count comparison of different configurations about the confidence map on ShanghaiTech Part\_A. X-axis: the ID of images, Y-axis: the crowd counts of images.}}
 
	\label{FIG:1}
\end{figure}

\begin{table*}
\setlength{\abovecaptionskip}{0.cm}
 
\setlength{\belowcaptionskip}{-0.9cm}

\caption{Estimation errors on the WorldExpo'10 dataset.}

\begin{center}
\scalebox{1.0}{
\label{my-label}
\begin{tabular}{ccccccc}
\hline
Method                                & S1 & S2 & S3 & S4 & S5 & Average \\ \hline
LBP+RR                                & 13.6   & 58.9  & 37.1  & 21.8  & 23.4   & 31.9 \\ 
SE Cycle GAN \cite{Wang_2019_CVPR}                   & 4.3   & 59.1  & 43.7  & 17.0  & 7.6   & 26.3 \\
MCNN \cite{zhang2016single}                   & 3.4   & 20.6  & 12.9  & 13.0  & 8.1   & 11.6 \\
TDF-CNN \cite{sam2018top}                & 2.7   & 23.4  & 10.7  & 17.6  & 3.3   & 11.5  \\
IG-CNN \cite{babu2018divide}                & 2.6   & 16.1  & 10.2   & 20.2  & 7.6   & 11.3  \\
Xiong. et al \cite{xiong2017spatiotemporal} & 6.8   & 14.5  & 14.9  & 13.5  & 3.1  & 10.6 \\
DDCN \cite{wang2019removing}       & 4.8   & 16.2  &12.4   & 10.9  & 4.9   & 9.8  \\   
Switch-CNN \cite{sam2017switching}        & 4.4   & 15.7  & 10.0  & 11.0 & 5.9   & 9.4  \\
DecideNet \cite{liu2018decidenet}       & 2.0   & 13.1  &\textbf{ 8.9}   & 17.4  & 4.8   & 9.2  \\
D-ConvNet-V1 \cite{shi2018crowd}       & \textbf{1.9}   & 12.1  & 20.7   & 8.3  & 2.6   & 9.1  \\ 
CP-CNN \cite{ sindagi2017generating}      & 2.9   & 14.7  & 10.5  &  10.4 & 5.8   & 8.9 \\ 
MRA-CNN \cite{zhang2019multi}                & 2.4   & 11.4  & 9.3  & 10.5  & 3.7   & 7.5  \\ 
ACSCP \cite{shen2018crowd}       & 2.8   & 14.1  & 9.6   &\textbf{ 8.1}  & 2.9   & 7.5  \\
CAT-CNN(OURS)                       & 2.2   & \textbf{9.8}  & 10.2  & 11.2  &\textbf{ 2.5}   &\textbf{7.2}  \\ \hline
\end{tabular}}
\end{center}
%\caption{Estimation errors on the WorldExpo'10 dataset.}
\end{table*}

\begin{table}[]
\setlength{\abovecaptionskip}{0.05cm}
\setlength{\belowcaptionskip}{-0.1cm}

\begin{center}
\centering
%\caption{Estimation errors on ShanghaiTech dataset}
\caption{Estimation errors on the ShanghaiTeach dataset.}
\scalebox{1.0}{
\label{my-label}
\begin{tabular}{cccccc}
\hline
Method                & \multicolumn{2}{c}{Part\_A}   & \multicolumn{2}{c}{Part\_B}  \\  \cmidrule(lr){2-3} \cmidrule(lr){4-5} 
                    & MAE           & MSE            & MAE           & MSE           \\ \hline

LBR + RR  & 303.2         & 371.0          & 59.1          & 81.7          \\ 
SE Cycle GAN \cite{Wang_2019_CVPR}        & 123.4         & 193.4         & 19.9         & 28.3          \\ 
MCNN \cite{zhang2016single}       & 110.2         & 173.2          & 26.4          & 41.3          \\ 
Switch-CNN \cite{sam2017switching} & 90.4          & 135.0          & 21.6          & 33.4          \\ 
TDF-CNN \cite{sam2018top}        & 97.5          & 145.1          & 20.7          & 32.8          \\ 
DecideNet+R3 \cite{liu2018decidenet}       &-         & -         & 20.8          & 29.4          \\ 

ACSCP \cite{shen2018crowd}   &75.7        &102.7        & 17.2 & 27.4            \\ 
MRA-CNN \cite{zhang2019multi}   &74.2        &112.5        & 11.9 & 21.3           \\
CP-CNN  \cite{sindagi2017generating}       &   73.6          & 106.4 & 20.1         & 30.1          \\
D-ConvNet-V1 \cite{shi2018crowd}   &73.5        &112.3        & 18.7 & 26.0            \\ 
Hossain et al. \cite{hossain2019crowd}       &-         & -         & 16.9          & 28.4          \\ 
IG-CNN \cite{babu2018divide}  &72.5        &118.2        & 13.6 & 21.2            \\ 
DDCN \cite{wang2019removing}  &71.5        &110.4        & 13.8 & 20.1            \\ 
CAT-CNN(OURS)              &\textbf{66.7}        &  \textbf{101.7 }       &\textbf{11.2}          &\textbf{20.0 }        \\ \hline
\end{tabular}}
\end{center}
%\caption{Estimation errors on the ShanghaiTeach dataset.}
\end{table}

\begin{table}[]
\setlength{\abovecaptionskip}{0.05cm}
\setlength{\belowcaptionskip}{-0.1cm}
%\captionsetup{belowskip=-15pt}
\begin{center}
\centering
%\caption{Estimation errors on ucf_cc_50 dataset}
\caption{Estimation errors on the UCF\_CC\_50 dataset.}
\scalebox{1.0}{
\label{my-label}
%\resizebox{150pt}{60pt}{
\begin{tabular}{ccccc}
\hline
\multicolumn{1}{c}{Method}                             & \multicolumn{1}{c}{MAE}            & \multicolumn{1}{c}{MSE}            \\ \hline
\multicolumn{1}{c}{Rodriguez et al.  \cite{rodriguez2011density} }         & \multicolumn{1}{c}{655.7}          & \multicolumn{1}{c}{697.8}          \\ 
\multicolumn{1}{c}{Lempitsky et al.     \cite{idrees2013multi} }        & \multicolumn{1}{c}{419.5}          & \multicolumn{1}{c}{541.6}          \\ 
\multicolumn{1}{c}{MCNN \cite{zhang2016single}}                & \multicolumn{1}{c}{377.6}          & \multicolumn{1}{c}{509.1}          \\ 
\multicolumn{1}{c}{SE Cycle GAN \cite{Wang_2019_CVPR}}                & \multicolumn{1}{c}{373.4}          & \multicolumn{1}{c}{528.8}          \\ 
\multicolumn{1}{c}{TDF-CNN \cite{sam2018top}}              & \multicolumn{1}{c}{354.7}          & \multicolumn{1}{c}{491.4} \\ 
\multicolumn{1}{c}{Switch-CNN \cite{sam2017switching}}              & \multicolumn{1}{c}{318.1}          & \multicolumn{1}{c}{439.2} \\ 
\multicolumn{1}{c}{CP-CNN \cite{sindagi2017generating}}              & \multicolumn{1}{c}{295.8}          & \multicolumn{1}{c}{320.9} \\ 
\multicolumn{1}{c}{IG-CNN \cite{babu2018divide}}              & \multicolumn{1}{c}{291.4}          & \multicolumn{1}{c}{349.4}          \\ 
\multicolumn{1}{c}{ACSCP \cite{shen2018crowd}}               & \multicolumn{1}{c}{291.0}          & \multicolumn{1}{c}{404.6}          \\ 
\multicolumn{1}{c}{D-ConvNet-V1 \cite{shi2018crowd}}        & \multicolumn{1}{c}{288.4}          & \multicolumn{1}{c}{404.7}          \\ 
\multicolumn{1}{c}{DDCN \cite{wang2019removing}}        & \multicolumn{1}{c}{286.2}          & \multicolumn{1}{c}{479.6}          \\ 
\multicolumn{1}{c}{Xiong. et al \cite{xiong2017spatiotemporal}} & \multicolumn{1}{c}{284.5}          & \multicolumn{1}{c}{\textbf{297.1}}          \\ 
\multicolumn{1}{c}{Hossain et al. \cite{hossain2019crowd}} & \multicolumn{1}{c}{271.6}          & \multicolumn{1}{c}{391.0}          \\ 
\multicolumn{1}{c}{MRA-CNN \cite{zhang2019multi}} & \multicolumn{1}{c}{240.8}          & \multicolumn{1}{c}{352.6}          \\
\multicolumn{1}{c}{CAT-CNN(OURS)}                            & \multicolumn{1}{c}{\textbf{235.5}} & \multicolumn{1}{c}{324.8}          \\ \hline      
\end{tabular}}
\end{center}
%\caption{Estimation errors on the UCF\_CC\_50 dataset.}

\end{table}

 %Before getting the exact crowd count, if the prior information of approximate total number of people in the image is used, the accuracy of crowd counting can be further enhanced. 

\subsection{Ablation study}

%We perform an ablation study on ShanghaiTech Part\_A \cite{zhang2016single}. Preliminary experiments were conducted to find the best configuration at the beginning of our CAT-CNN, three types of configurations about different kernel sizes are listed in Table 1, and it can be found that using four different kernel sizes is the best. In order to demonstrate the crowd count group classifier is helpful to reduce the crowd counting errors, in the comparative experiment, the FC layers in the  Multi-information Handling Module  is removed.  In order to demonstrate the effects of Confidence Module, we also conduct a set of comparative experiments. In one experiment, the Confidence Module can function normally. In other experiment, the Confidence Module is removed. The results of the comparative experiments are shown in Fig. 5. The count estimation errors of different configurations are illustrated in Table 1. From these results we can see that the proposed CAT-CNN can further reduce the estimation errors by using suitable receptive fields information and attention information.

To further demonstrate the effectiveness of different components in our CAT-CNN, we perform an ablation study by a discussion. 
%\textcolor{red}{on ShanghaiTech Part\_A \cite{zhang2016single} which contains the appropriate number of images, the diverse crowd distribution and the maximum number of labeled people.} Similar results can be found in other datasets.

\noindent{\bfseries Benefits of Multi-Scale Features:} The crowd at different distances from the camera have different scale characteristics in the image. So it is very important to extract multi-scale features. In \cite{zhang2016single},  Zhang et al. proposed the three-column CNN structures to extract multi-scale features. However, every column needs to be pre-trained separately and the multi-column CNNs are hard to be trained. By contrast, our model is more convenient and effective to extract multi-scale features.

As shown in Fig. 3, at the beginning of our model, the convolution kernels with different scales are used to map the head to feature maps from input image. From Table 3, it can be found that the performance of using four type convolution kernels is best in most cases. As shown in Fig. 3, inside the network, feature maps from convolutional layers of different depths are merged by cross-layer connection to automatically adapt the scale variations in the crowd. Higher layers encode the semantic concept of the person, whereas lower layers extract rich discriminative features from the person. Both of them can provide complementary information on the same person with different levels. Hence, from Table 4, we can observe that fusing features from convolutional layers of different depths is very effective to alleviate the crowd counting errors on these benchmarking datasets. %with the MAE/MSE 8.8/23.2 lower than that without cross-layer connection on ShanghaiTech Part\_A.

\noindent{\bfseries Benefits of The Prior of Population-Level Categories:} In our CAT-CNN, we classify the crowd counts of images in each dataset into 5 categories according to experiments. The results are shown in Table 2.  When the number of categories is set to 15 on the WorldExpo’10 dataset, the performance is improved slightly compared with the third column. We think that due to a large number of images in this dataset, 15 categories are suitable. However, it can be observed that the performance of 5 categories is best in most cases. We argue that due to the limitations of model capacity and training data, 5 categories are sufficient for most datasets. To prove the effectiveness of explicit use of the prior of population-level category, in the comparative experiment, FM1 and FM2 respectively serve as the output of the Multi-information Handling Module. The results are shown in Table 5. We can observe that by only using FM2 where we explicitly use the prior of population-level categories, the performance is better, with the MAE/MSE 5.4/6.2 lower than only using FM1 where we do not explicitly use the prior of population-level categories. When FM1 and FM2 are concatenated and used simultaneously, the performance is the best, with the MAE/MSE 10.0/13.6 lower than only using FM1. We argue that both FM1 and FM2 can encode the crowd count feature. The concatenation of them can provide more sufficient feature information for following modules.

\noindent{\bfseries Benefits of The Confidence Map:} To demonstrate the effectiveness of the confidence map, performances of our model with and without confidence map are compared. In the comparative experiment, the Confidence Module is removed from our CAT-CNN. Comparison results are given in Table 6. We can see that the performances of our model are further enhanced by a confidence map generated in the Confidence Module, with the MAE/MSE 14.5/27.2 lower than that without a confidence map. The estimated count of each image and its actual count are shown in Fig. 10. We can see that the green line (estimated count with confidence map) is closer to the red line (actual count) than the blue line (estimated count without confidence map), which denotes that the confidence map plays an important role in reducing the error of crowd counting.

 To visualize the effectiveness of confidence map, the confidence map is resized to the same size as the input image. And it is overlaid on the input image with 70\% transparency to display both of them clearly. They are shown in the second row of Fig. 8. We can observe that only the position of the human head is highlighted, which denotes that our CAT-CNN encodes an accurate confidence map to avoid enormous misjudgements. The ground-truth density map and final high-precision density map are also shown in Fig. 8. We can observe that our CAT-CNN encodes a high-precision density map to count the crowd very well. Similar results can be found in Fig. 9.

\noindent{\bfseries Evaluations of The Real Time:} To verify the real-time performance of our proposed method, some experiments are conducted on a video surveillance dataset (WorldExpo’10 dataset) using NVIDIA GTX 1080Ti and Intel Core i7. Its test set contains 600 frames from 5 different scenes (S1-S5). The resolution of each frame is 576×720. Comparisons with other state-of-the-art methods are given in Table 7. It can be observed that our proposed method obtains competitive results.

% and Fig. 9
%To visualize the effect of confidence map, the confidence map is resized to the same size of original image. And it is overlaied on the original image with 70\% transparency to display both of them clearly. The results are shown in Fig. 5. We can see that our CAT-CNN with the guidance of the confidence map can avoid misjudgment very well. The estimated count of each image and its actual count are shown in Fig. 5. We can see that the estimated count with confidence map (red line) is closer the actual count (green line), which denotes that the confidence map plays a crucial role in crowd counting. 
%is helpful to estimate the crowd counts accurately.

\subsection{Shanghaitech dataset}   
Zhang et al. \cite{zhang2016single} introduced the large-scale ShanghaiTech dataset. It consists of two parts: Part\_A with 300 training images and 182 test images, Part\_B with 400 training images and 316 test images. In Part\_A, the crowd density is high while the crowd density is relatively low in Part\_B. We  generate the ground-truth density map by using the method in Sec. 3.2.

%The estimated count and its actual count of each image are shown in Fig. 10. We can see that the green line (estimated count by our CAT-CNN) follows the red line (actual count) tightly, which denotes that our CAT-CNN can count the crowd with a small error in most images. We can also observe that the crowd count error is relatively large in the image with extremely dense crowds. We think that lack of dense crowd training data leads to this result.       
%As shown in Table 3, our proposed method is compared with other state-of-the-art methods. We can see that whether in sparse crowds or dense crowds, our CAT-CNN still outperform other methods.
  
  In Table 9, we compare our method with other recent state-of-the-art methods. The LBP and RR are traditional algorithms for crowd counting. In \cite{Wang_2019_CVPR}, the data collector and labeler were proposed to generate and annotate the crowd data. In \cite{zhang2016single}, the MCNN with three-column CNNs was proposed to extract multi-scale features to adapt scale variations. In \cite{sam2017switching}, the Switch-CNN with a classifier was proposed to select an optimal branch to generate the density map. In \cite{sindagi2017generating}, global and local features were used to generate a high-quality density map.  In \cite{sam2018top}, the top-down feedback TDF-CNN was proposed to get initial accurate prediction.  In \cite{liu2018decidenet}, detection and regression were used for crowd counting simultaneously. In \cite{shen2018crowd}, the GANs were proposed to mitigate blurring in the estimated density map. In \cite{zhang2019multi}, head regions were focused on automatically. In \cite{shi2018crowd}, the deep negative correlation learning was proposed to reduce over-fitting. In \cite{hossain2019crowd}, a scale-aware attention model was proposed to adapt the scale variation of crowds. In \cite{babu2018divide}, a growing CNN was proposed to adapt the variability seen from the crowd by increasing its capacity. In \cite{wang2019removing}, the interference from background could be removed to automatically alleviate the mapping between input images and crowd counts. We can find that most of them pay attention to the extraction of multi-scale features in the crowd. Because the scale variation of crowds restricts the performance of proposed methods.
 
In Table 9, it can be observed that all of the CNN-based methods have an absolute advantage over the traditional algorithms. In the CNN-based methods, our method achieves the best results on both Part\_A and Part\_B, which indicates that the accurate judgement of human head is important for improving the accuracy of crowd counting. We can also observe that Part\_A is more challenging than Part\_B. The crowd density is higher and the training is harder. As shown in Fig. 11 that is from Part\_A, the final estimated density map and its ground truth are very similar, except for the region in red rectangles where people are more difficult to recognize and these samples are more difficult to train. In the future, we plan to introduce the hard example mining technology to break through the limitation to further improve counting accuracy.

 %The methods \cite{zhang2016single,sam2018top,sam2017switching,sindagi2017generating,babu2018divide,shen2018crowd,shi2018crowd,liu2018leveraging} on the Shanghaitech dataset are still compared on UCF\_CC\_50 dataset.

%\cite{zhang2016single,sam2017switching,sindagi2017generating,sam2018top,liu2018decidenet,shen2018crowd,shi2018crowd,babu2018dividev,liu2018leveraging}

\subsection{UCF\_CC\_50 dataset} 
Idrees et al. \cite{idrees2013multi} collected 50 images from internet to produce the challenging UCF\_CC\_50 dataset. The number of annotated heads in each image ranges from 94 to 4,543. The total number of people in the dataset is 63,974.  It is a challenging dataset because of its dense crowd, limited image, and low resolution. We generate the ground-truth density map by using the method in Sec. 3.2. The 5-fold cross validation is performed to evaluate proposed methods.

 %In \cite{amirgholipour2018ccnn}, the fuzzy inference engine was proposed to adapt changes in head size. In \cite{ranjan2018iterative}, a two-branch CNN model was proposed to generate the high-resolution density map.
%In \cite{rodriguez2011density,idrees2013multi}, the traditional feature extractors such as HOG and SIFT are used to obtain the crowd counts. Other methods are based on deep learning. The methods \cite{zhang2016single,sam2017switching,sam2018top,shen2018crowd,zhang2019multi,sindagi2017generating,shi2018crowd,babu2018divide,wang2019removing,liu2018leveraging} on the Shanghaitech dataset are still compared on UCF\_CC\_50 dataset. 
In Table 10, Other recent state-of-the-art methods are compared with our method. In \cite{rodriguez2011density,idrees2013multi}, traditional feature extractors were used to extract crowd features. In \cite{xiong2017spatiotemporal}, the ConvLSTM employed the temporal correlation to assist crowd counting. Other methods on Shanghaitech dataset are still compared on this dataset. In Table 10, we can observe that the density map learned by CNN is more robust than the hand-crafted features extracted in  \cite{rodriguez2011density,idrees2013multi}. Our method outperforms all other methods on MAE metric, while we obtain a competitive MSE score which denotes the robustness of proposed methods. We argue that the limited data for training cause this result as the UCF\_CC\_50 dataset only contains 50 images.

\subsection{WorldExpo’10 dataset} 

The WorldExpo’10 dataset \cite{zhang2015cross} consists of 3,980 annotated frames collected from 1,132 video sequences. The video sequences contain 108 different scenes. This dataset is divided into a training set with 3,380 frames and a test set with 600 frames from 5 different scenes (S1-S5). This dataset provides the region of interest (ROI) and perspective map for each scene. For fair comparisons, we follow the experimental setting in \cite{zhang2015cross} to generate density maps.

In Table 8, we compare our method with other recent state-of-the-art methods. MAE is used to evaluate the results of different methods, which is suggested in \cite{zhang2015cross}. We can observe that our proposed method obtains the best results on the average MAE of five different scenes. It can be also observed that we get competitive results in the three of five scenes. By reviewing all test images, we find that due to the effect of ROI, the images in these three scenes are no longer complicated. There are fewer people and little background noise. Therefore, it is difficult to play our strengths.
%However, our method only reveals 0.2 improvement on the second best method whose average MAE is 9.2. The reason may lie on Sce.3 with a little higher MAE. In Sce.3, some people are severely obstructed by the tent, which is a great challenge for our CAT-CNN.   

\begin{figure}

   \centering
   \includegraphics[width=1.0\linewidth]{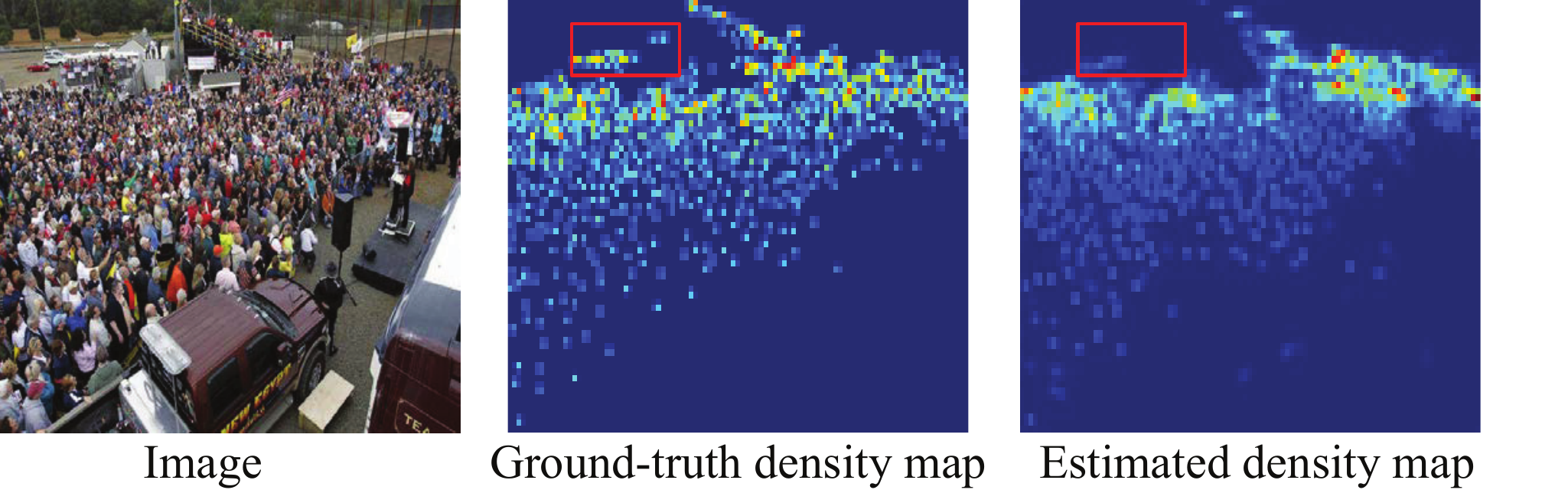}

   %\centering{\caption{The proposed architecture of our CAT-CNN}*}
   \centering{\caption{Limitation of the proposed method}}
 
	\label{FIG:1}
\end{figure}

\section{Conclusion}
\label{sec:conclusion}

 In this paper, we proposed an end-to-end model named CAT-CNN in crowd counting. Our CAT-CNN can adaptively assess the importance of a human head at each pixel location to avoid enormous misjudgements. To obtain a robust confidence map for population distribution, the weighted BCELoss is proposed. The crowd counts are classified into five groups in each dataset and we first explicitly map the prior of the population-level category to feature maps to automatically contribute in encoding a highly refined density map. We evaluate our method on three benchmark datasets separately. Extensive experimental results indicate that our method outperforms many state-of-the-art methods. In the future, we will focus on hard example mining technology to handle the problem proposed in Fig. 11.

\noindent{\bfseries Acknowledgments.} This work was supported by the National Natural Science Foundation of China (No.61472393).

\noindent{\bfseries \textcolor{red}{Please cite
@article\{CHEN2020,
  title=\{Crowd counting with crowd attention convolutional neural network\},
  author=\{ Chen, Jiwei  and  Wen, Su  and  Wang, Zengfu\},
  journal=\{Neurocomputing\},
  volume=\{382\},
  pages=\{210--220\},
  year=\{2020\},
  publisher=\{Elsevier\}
\}}
}

%\section{Acknowledgments}
%This work was supported by the National Natural Science Foundation of China (No.61472393).

\bibliographystyle{cas-model2-names}

\bibliography{mulu_13_1}

\begin{thebibliography}{46}
\expandafter\ifx\csname natexlab\endcsname\relax\def\natexlab#1{#1}\fi
\providecommand{\url}[1]{\texttt{#1}}
\providecommand{\href}[2]{#2}
\providecommand{\path}[1]{#1}
\providecommand{\DOIprefix}{doi:}
\providecommand{\ArXivprefix}{arXiv:}
\providecommand{\URLprefix}{URL: }
\providecommand{\Pubmedprefix}{pmid:}
\providecommand{\doi}[1]{\href{http://dx.doi.org/#1}{\path{#1}}}
\providecommand{\Pubmed}[1]{\href{pmid:#1}{\path{#1}}}
\providecommand{\bibinfo}[2]{#2}
\ifx\xfnm\relax \def\xfnm[#1]{\unskip,\space#1}\fi
%Type = Inproceedings
\bibitem[{Babu~Sam et~al.(2018)Babu~Sam, Sajjan, Venkatesh~Babu and
  Srinivasan}]{babu2018divide}
\bibinfo{author}{Babu~Sam, D.}, \bibinfo{author}{Sajjan, N.N.},
  \bibinfo{author}{Venkatesh~Babu, R.}, \bibinfo{author}{Srinivasan, M.},
  \bibinfo{year}{2018}.
\newblock \bibinfo{title}{Divide and grow: Capturing huge diversity in crowd
  images with incrementally growing cnn}, in: \bibinfo{booktitle}{Proceedings
  of the IEEE Conference on Computer Vision and Pattern Recognition}, pp.
  \bibinfo{pages}{3618--3626}.
%Type = Inproceedings
\bibitem[{Boominathan et~al.(2016)Boominathan, Kruthiventi and
  Babu}]{boominathan2016crowdnet}
\bibinfo{author}{Boominathan, L.}, \bibinfo{author}{Kruthiventi, S.S.},
  \bibinfo{author}{Babu, R.V.}, \bibinfo{year}{2016}.
\newblock \bibinfo{title}{Crowdnet: A deep convolutional network for dense
  crowd counting}, in: \bibinfo{booktitle}{Proceedings of the 24th ACM
  international conference on Multimedia}, \bibinfo{organization}{ACM}. pp.
  \bibinfo{pages}{640--644}.
%Type = Inproceedings
\bibitem[{Chan et~al.(2008)Chan, Liang and Vasconcelos}]{chan2008privacy}
\bibinfo{author}{Chan, A.B.}, \bibinfo{author}{Liang, Z.S.J.},
  \bibinfo{author}{Vasconcelos, N.}, \bibinfo{year}{2008}.
\newblock \bibinfo{title}{Privacy preserving crowd monitoring: Counting people
  without people models or tracking}, in: \bibinfo{booktitle}{2008 IEEE
  Conference on Computer Vision and Pattern Recognition},
  \bibinfo{organization}{IEEE}. pp. \bibinfo{pages}{1--7}.
%Type = Inproceedings
\bibitem[{Chan and Vasconcelos(2009)}]{chan2009bayesian}
\bibinfo{author}{Chan, A.B.}, \bibinfo{author}{Vasconcelos, N.},
  \bibinfo{year}{2009}.
\newblock \bibinfo{title}{Bayesian poisson regression for crowd counting}, in:
  \bibinfo{booktitle}{Computer Vision, 2009 IEEE 12th International Conference
  on}, \bibinfo{organization}{IEEE}. pp. \bibinfo{pages}{545--551}.
%Type = Inproceedings
\bibitem[{Collobert and Weston(2008)}]{collobert2008unified}
\bibinfo{author}{Collobert, R.}, \bibinfo{author}{Weston, J.},
  \bibinfo{year}{2008}.
\newblock \bibinfo{title}{A unified architecture for natural language
  processing: Deep neural networks with multitask learning}, in:
  \bibinfo{booktitle}{Proceedings of the 25th international conference on
  Machine learning}, \bibinfo{organization}{ACM}. pp.
  \bibinfo{pages}{160--167}.
%Type = Inproceedings
\bibitem[{Dalal and Triggs(2005)}]{dalal2005histograms}
\bibinfo{author}{Dalal, N.}, \bibinfo{author}{Triggs, B.},
  \bibinfo{year}{2005}.
\newblock \bibinfo{title}{Histograms of oriented gradients for human
  detection}, in: \bibinfo{booktitle}{Computer Vision and Pattern Recognition,
  2005. CVPR 2005. IEEE Computer Society Conference on},
  \bibinfo{organization}{IEEE}. pp. \bibinfo{pages}{886--893}.
%Type = Inproceedings
\bibitem[{Glorot et~al.(2011)Glorot, Bordes and Bengio}]{glorot2011deep}
\bibinfo{author}{Glorot, X.}, \bibinfo{author}{Bordes, A.},
  \bibinfo{author}{Bengio, Y.}, \bibinfo{year}{2011}.
\newblock \bibinfo{title}{Deep sparse rectifier neural networks}, in:
  \bibinfo{booktitle}{Proceedings of the fourteenth international conference on
  artificial intelligence and statistics}, pp. \bibinfo{pages}{315--323}.
%Type = Inproceedings
\bibitem[{He et~al.(2015)He, Zhang, Ren and Sun}]{he2015delving}
\bibinfo{author}{He, K.}, \bibinfo{author}{Zhang, X.}, \bibinfo{author}{Ren,
  S.}, \bibinfo{author}{Sun, J.}, \bibinfo{year}{2015}.
\newblock \bibinfo{title}{Delving deep into rectifiers: Surpassing human-level
  performance on imagenet classification}, in: \bibinfo{booktitle}{Proceedings
  of the IEEE international conference on computer vision}, pp.
  \bibinfo{pages}{1026--1034}.
%Type = Inproceedings
\bibitem[{He et~al.(2016)He, Zhang, Ren and Sun}]{he2016deep}
\bibinfo{author}{He, K.}, \bibinfo{author}{Zhang, X.}, \bibinfo{author}{Ren,
  S.}, \bibinfo{author}{Sun, J.}, \bibinfo{year}{2016}.
\newblock \bibinfo{title}{Deep residual learning for image recognition}, in:
  \bibinfo{booktitle}{Proceedings of the IEEE conference on computer vision and
  pattern recognition}, pp. \bibinfo{pages}{770--778}.
%Type = Inproceedings
\bibitem[{Hossain et~al.(2019)Hossain, Hosseinzadeh, Chanda and
  Wang}]{hossain2019crowd}
\bibinfo{author}{Hossain, M.}, \bibinfo{author}{Hosseinzadeh, M.},
  \bibinfo{author}{Chanda, O.}, \bibinfo{author}{Wang, Y.},
  \bibinfo{year}{2019}.
\newblock \bibinfo{title}{Crowd counting using scale-aware attention networks},
  in: \bibinfo{booktitle}{2019 IEEE Winter Conference on Applications of
  Computer Vision (WACV)}, \bibinfo{organization}{IEEE}. pp.
  \bibinfo{pages}{1280--1288}.
%Type = Article
\bibitem[{Hu et~al.(2016)Hu, Chang, Nian, Wang and Li}]{hu2016dense}
\bibinfo{author}{Hu, Y.}, \bibinfo{author}{Chang, H.}, \bibinfo{author}{Nian,
  F.}, \bibinfo{author}{Wang, Y.}, \bibinfo{author}{Li, T.},
  \bibinfo{year}{2016}.
\newblock \bibinfo{title}{Dense crowd counting from still images with
  convolutional neural networks}.
\newblock \bibinfo{journal}{Journal of Visual Communication and Image
  Representation} \bibinfo{volume}{38}, \bibinfo{pages}{530--539}.
%Type = Inproceedings
\bibitem[{Idrees et~al.(2013)Idrees, Saleemi, Seibert and
  Shah}]{idrees2013multi}
\bibinfo{author}{Idrees, H.}, \bibinfo{author}{Saleemi, I.},
  \bibinfo{author}{Seibert, C.}, \bibinfo{author}{Shah, M.},
  \bibinfo{year}{2013}.
\newblock \bibinfo{title}{Multi-source multi-scale counting in extremely dense
  crowd images}, in: \bibinfo{booktitle}{Proceedings of the IEEE conference on
  computer vision and pattern recognition}, pp. \bibinfo{pages}{2547--2554}.
%Type = Inproceedings
\bibitem[{Li et~al.(2008)Li, Zhang, Huang and Tan}]{li2008estimating}
\bibinfo{author}{Li, M.}, \bibinfo{author}{Zhang, Z.}, \bibinfo{author}{Huang,
  K.}, \bibinfo{author}{Tan, T.}, \bibinfo{year}{2008}.
\newblock \bibinfo{title}{Estimating the number of people in crowded scenes by
  mid based foreground segmentation and head-shoulder detection}, in:
  \bibinfo{booktitle}{2008 19th International Conference on Pattern
  Recognition}, \bibinfo{organization}{IEEE}. pp. \bibinfo{pages}{1--4}.
%Type = Article
\bibitem[{Li et~al.(2015)Li, Chang, Wang, Ni, Hong and Yan}]{li2015crowded}
\bibinfo{author}{Li, T.}, \bibinfo{author}{Chang, H.}, \bibinfo{author}{Wang,
  M.}, \bibinfo{author}{Ni, B.}, \bibinfo{author}{Hong, R.},
  \bibinfo{author}{Yan, S.}, \bibinfo{year}{2015}.
\newblock \bibinfo{title}{Crowded scene analysis: A survey}.
\newblock \bibinfo{journal}{IEEE transactions on circuits and systems for video
  technology} \bibinfo{volume}{25}, \bibinfo{pages}{367--386}.
%Type = Article
\bibitem[{Li et~al.(2018)Li, Wang, Hong, Wang and Wu}]{li2018pdisvpl}
\bibinfo{author}{Li, T.}, \bibinfo{author}{Wang, Y.}, \bibinfo{author}{Hong,
  R.}, \bibinfo{author}{Wang, M.}, \bibinfo{author}{Wu, X.},
  \bibinfo{year}{2018}.
\newblock \bibinfo{title}{pdisvpl: Probabilistic discriminative visual part
  learning for image classification}.
\newblock \bibinfo{journal}{IEEE MultiMedia} \bibinfo{volume}{25},
  \bibinfo{pages}{34--45}.
%Type = Article
\bibitem[{Lin et~al.(2013)Lin, Chen and Yan}]{lin2013network}
\bibinfo{author}{Lin, M.}, \bibinfo{author}{Chen, Q.}, \bibinfo{author}{Yan,
  S.}, \bibinfo{year}{2013}.
\newblock \bibinfo{title}{Network in network}.
\newblock \bibinfo{journal}{arXiv preprint arXiv:1312.4400} .
%Type = Article
\bibitem[{Lin and Davis(2010)}]{lin2010shape}
\bibinfo{author}{Lin, Z.}, \bibinfo{author}{Davis, L.S.}, \bibinfo{year}{2010}.
\newblock \bibinfo{title}{Shape-based human detection and segmentation via
  hierarchical part-template matching}.
\newblock \bibinfo{journal}{IEEE Transactions on Pattern Analysis and Machine
  Intelligence} \bibinfo{volume}{32}, \bibinfo{pages}{604--618}.
%Type = Inproceedings
\bibitem[{Liu et~al.(2018)Liu, Gao, Meng and Hauptmann}]{liu2018decidenet}
\bibinfo{author}{Liu, J.}, \bibinfo{author}{Gao, C.}, \bibinfo{author}{Meng,
  D.}, \bibinfo{author}{Hauptmann, A.G.}, \bibinfo{year}{2018}.
\newblock \bibinfo{title}{Decidenet: Counting varying density crowds through
  attention guided detection and density estimation}, in:
  \bibinfo{booktitle}{Proceedings of the IEEE Conference on Computer Vision and
  Pattern Recognition}, pp. \bibinfo{pages}{5197--5206}.
%Type = Inproceedings
\bibitem[{Long et~al.(2015)Long, Shelhamer and Darrell}]{long2015fully}
\bibinfo{author}{Long, J.}, \bibinfo{author}{Shelhamer, E.},
  \bibinfo{author}{Darrell, T.}, \bibinfo{year}{2015}.
\newblock \bibinfo{title}{Fully convolutional networks for semantic
  segmentation}, in: \bibinfo{booktitle}{Proceedings of the IEEE conference on
  computer vision and pattern recognition}, pp. \bibinfo{pages}{3431--3440}.
%Type = Inproceedings
\bibitem[{Marana et~al.(1998)Marana, Costa, Lotufo and
  Velastin}]{marana1998efficacy}
\bibinfo{author}{Marana, A.}, \bibinfo{author}{Costa, L.d.F.},
  \bibinfo{author}{Lotufo, R.}, \bibinfo{author}{Velastin, S.},
  \bibinfo{year}{1998}.
\newblock \bibinfo{title}{On the efficacy of texture analysis for crowd
  monitoring}, in: \bibinfo{booktitle}{Proceedings SIBGRAPI'98. International
  Symposium on Computer Graphics, Image Processing, and Vision (Cat. No.
  98EX237)}, \bibinfo{organization}{IEEE}. pp. \bibinfo{pages}{354--361}.
%Type = Article
\bibitem[{Marsde et~al.(2018)Marsde, McGuinness, Little, Keogh and
  O'Connor}]{marsde2018people}
\bibinfo{author}{Marsde, M.}, \bibinfo{author}{McGuinness, K.},
  \bibinfo{author}{Little, S.}, \bibinfo{author}{Keogh, C.E.},
  \bibinfo{author}{O'Connor, N.E.}, \bibinfo{year}{2018}.
\newblock \bibinfo{title}{People, penguins and petri dishes: adapting object
  counting models to new visual domains and object types without forgetting} .
%Type = Inproceedings
\bibitem[{Ren et~al.(2015)Ren, He, Girshick and Sun}]{ren2015faster}
\bibinfo{author}{Ren, S.}, \bibinfo{author}{He, K.}, \bibinfo{author}{Girshick,
  R.}, \bibinfo{author}{Sun, J.}, \bibinfo{year}{2015}.
\newblock \bibinfo{title}{Faster r-cnn: Towards real-time object detection with
  region proposal networks}, in: \bibinfo{booktitle}{Advances in neural
  information processing systems}, pp. \bibinfo{pages}{91--99}.
%Type = Inproceedings
\bibitem[{Rodriguez et~al.(2011)Rodriguez, Laptev, Sivic and
  Audibert}]{rodriguez2011density}
\bibinfo{author}{Rodriguez, M.}, \bibinfo{author}{Laptev, I.},
  \bibinfo{author}{Sivic, J.}, \bibinfo{author}{Audibert, J.Y.},
  \bibinfo{year}{2011}.
\newblock \bibinfo{title}{Density-aware person detection and tracking in
  crowds}, in: \bibinfo{booktitle}{2011 International Conference on Computer
  Vision}, \bibinfo{organization}{IEEE}. pp. \bibinfo{pages}{2423--2430}.
%Type = Inproceedings
\bibitem[{Ryan et~al.(2009)Ryan, Denman, Fookes and Sridharan}]{ryan2009crowd}
\bibinfo{author}{Ryan, D.}, \bibinfo{author}{Denman, S.},
  \bibinfo{author}{Fookes, C.}, \bibinfo{author}{Sridharan, S.},
  \bibinfo{year}{2009}.
\newblock \bibinfo{title}{Crowd counting using multiple local features}, in:
  \bibinfo{booktitle}{2009 Digital Image Computing: Techniques and
  Applications}, \bibinfo{organization}{IEEE}. pp. \bibinfo{pages}{81--88}.
%Type = Inproceedings
\bibitem[{Sam and Babu(2018)}]{sam2018top}
\bibinfo{author}{Sam, D.B.}, \bibinfo{author}{Babu, R.V.},
  \bibinfo{year}{2018}.
\newblock \bibinfo{title}{Top-down feedback for crowd counting convolutional
  neural network}, in: \bibinfo{booktitle}{Thirty-Second AAAI Conference on
  Artificial Intelligence}.
%Type = Inproceedings
\bibitem[{Sam et~al.(2017)Sam, Surya and Babu}]{sam2017switching}
\bibinfo{author}{Sam, D.B.}, \bibinfo{author}{Surya, S.},
  \bibinfo{author}{Babu, R.V.}, \bibinfo{year}{2017}.
\newblock \bibinfo{title}{Switching convolutional neural network for crowd
  counting}, in: \bibinfo{booktitle}{Proceedings of the IEEE Conference on
  Computer Vision and Pattern Recognition}, p.~\bibinfo{pages}{6}.
%Type = Inproceedings
\bibitem[{Shen et~al.(2018)Shen, Xu, Ni, Wang, Hu and Yang}]{shen2018crowd}
\bibinfo{author}{Shen, Z.}, \bibinfo{author}{Xu, Y.}, \bibinfo{author}{Ni, B.},
  \bibinfo{author}{Wang, M.}, \bibinfo{author}{Hu, J.}, \bibinfo{author}{Yang,
  X.}, \bibinfo{year}{2018}.
\newblock \bibinfo{title}{Crowd counting via adversarial cross-scale
  consistency pursuit}, in: \bibinfo{booktitle}{Proceedings of the IEEE
  Conference on Computer Vision and Pattern Recognition}, pp.
  \bibinfo{pages}{5245--5254}.
%Type = Inproceedings
\bibitem[{Shi et~al.(2018)Shi, Zhang, Liu, Cao, Ye, Cheng and
  Zheng}]{shi2018crowd}
\bibinfo{author}{Shi, Z.}, \bibinfo{author}{Zhang, L.}, \bibinfo{author}{Liu,
  Y.}, \bibinfo{author}{Cao, X.}, \bibinfo{author}{Ye, Y.},
  \bibinfo{author}{Cheng, M.M.}, \bibinfo{author}{Zheng, G.},
  \bibinfo{year}{2018}.
\newblock \bibinfo{title}{Crowd counting with deep negative correlation
  learning}, in: \bibinfo{booktitle}{Proceedings of the IEEE Conference on
  Computer Vision and Pattern Recognition}, pp. \bibinfo{pages}{5382--5390}.
%Type = Inproceedings
\bibitem[{Sindagi and Patel(2017a)}]{sindagi2017cnn}
\bibinfo{author}{Sindagi, V.A.}, \bibinfo{author}{Patel, V.M.},
  \bibinfo{year}{2017}a.
\newblock \bibinfo{title}{Cnn-based cascaded multi-task learning of high-level
  prior and density estimation for crowd counting}, in:
  \bibinfo{booktitle}{2017 14th IEEE International Conference on Advanced Video
  and Signal Based Surveillance (AVSS)}, \bibinfo{organization}{IEEE}. pp.
  \bibinfo{pages}{1--6}.
%Type = Inproceedings
\bibitem[{Sindagi and Patel(2017b)}]{sindagi2017generating}
\bibinfo{author}{Sindagi, V.A.}, \bibinfo{author}{Patel, V.M.},
  \bibinfo{year}{2017}b.
\newblock \bibinfo{title}{Generating high-quality crowd density maps using
  contextual pyramid cnns}, in: \bibinfo{booktitle}{2017 IEEE International
  Conference on Computer Vision (ICCV)}, \bibinfo{organization}{IEEE}. pp.
  \bibinfo{pages}{1879--1888}.
%Type = Article
\bibitem[{Sindagi and Patel(2018)}]{sindagi2018survey}
\bibinfo{author}{Sindagi, V.A.}, \bibinfo{author}{Patel, V.M.},
  \bibinfo{year}{2018}.
\newblock \bibinfo{title}{A survey of recent advances in cnn-based single image
  crowd counting and density estimation}.
\newblock \bibinfo{journal}{Pattern Recognition Letters} \bibinfo{volume}{107},
  \bibinfo{pages}{3--16}.
%Type = Inproceedings
\bibitem[{Szegedy et~al.(2015)Szegedy, Liu, Jia, Sermanet, Reed, Anguelov,
  Erhan, Vanhoucke and Rabinovich}]{szegedy2015going}
\bibinfo{author}{Szegedy, C.}, \bibinfo{author}{Liu, W.}, \bibinfo{author}{Jia,
  Y.}, \bibinfo{author}{Sermanet, P.}, \bibinfo{author}{Reed, S.},
  \bibinfo{author}{Anguelov, D.}, \bibinfo{author}{Erhan, D.},
  \bibinfo{author}{Vanhoucke, V.}, \bibinfo{author}{Rabinovich, A.},
  \bibinfo{year}{2015}.
\newblock \bibinfo{title}{Going deeper with convolutions}, in:
  \bibinfo{booktitle}{Proceedings of the IEEE conference on computer vision and
  pattern recognition}, pp. \bibinfo{pages}{1--9}.
%Type = Article
\bibitem[{Viola and Jones(2004)}]{viola2004robust}
\bibinfo{author}{Viola, P.}, \bibinfo{author}{Jones, M.J.},
  \bibinfo{year}{2004}.
\newblock \bibinfo{title}{Robust real-time face detection}.
\newblock \bibinfo{journal}{International journal of computer vision}
  \bibinfo{volume}{57}, \bibinfo{pages}{137--154}.
%Type = Article
\bibitem[{Wang et~al.(2016)Wang, Sun, Zhang, Thomas, Duan and
  Shi}]{wang2016large}
\bibinfo{author}{Wang, J.}, \bibinfo{author}{Sun, Y.}, \bibinfo{author}{Zhang,
  W.}, \bibinfo{author}{Thomas, I.}, \bibinfo{author}{Duan, S.},
  \bibinfo{author}{Shi, Y.}, \bibinfo{year}{2016}.
\newblock \bibinfo{title}{Large-scale online multitask learning and decision
  making for flexible manufacturing}.
\newblock \bibinfo{journal}{IEEE Transactions on Industrial Informatics}
  \bibinfo{volume}{12}, \bibinfo{pages}{2139--2147}.
%Type = Article
\bibitem[{Wang et~al.(2019a)Wang, Yin, Tang and Li}]{wang2019removing}
\bibinfo{author}{Wang, L.}, \bibinfo{author}{Yin, B.}, \bibinfo{author}{Tang,
  X.}, \bibinfo{author}{Li, Y.}, \bibinfo{year}{2019}a.
\newblock \bibinfo{title}{Removing background interference for crowd counting
  via de-background detail convolutional network}.
\newblock \bibinfo{journal}{Neurocomputing} \bibinfo{volume}{332},
  \bibinfo{pages}{360--371}.
%Type = Inproceedings
\bibitem[{Wang et~al.(2019b)Wang, Gao, Lin and Yuan}]{Wang_2019_CVPR}
\bibinfo{author}{Wang, Q.}, \bibinfo{author}{Gao, J.}, \bibinfo{author}{Lin,
  W.}, \bibinfo{author}{Yuan, Y.}, \bibinfo{year}{2019}b.
\newblock \bibinfo{title}{Learning from synthetic data for crowd counting in
  the wild}, in: \bibinfo{booktitle}{The IEEE Conference on Computer Vision and
  Pattern Recognition (CVPR)}.
%Type = Inproceedings
\bibitem[{Wang et~al.(2009)Wang, Han and Yan}]{wang2009hog}
\bibinfo{author}{Wang, X.}, \bibinfo{author}{Han, T.X.}, \bibinfo{author}{Yan,
  S.}, \bibinfo{year}{2009}.
\newblock \bibinfo{title}{An hog-lbp human detector with partial occlusion
  handling}, in: \bibinfo{booktitle}{2009 IEEE 12th international conference on
  computer vision}, \bibinfo{organization}{IEEE}. pp. \bibinfo{pages}{32--39}.
%Type = Inproceedings
\bibitem[{Xiong et~al.(2017)Xiong, Shi and Yeung}]{xiong2017spatiotemporal}
\bibinfo{author}{Xiong, F.}, \bibinfo{author}{Shi, X.}, \bibinfo{author}{Yeung,
  D.Y.}, \bibinfo{year}{2017}.
\newblock \bibinfo{title}{Spatiotemporal modeling for crowd counting in
  videos}, in: \bibinfo{booktitle}{2017 IEEE International Conference on
  Computer Vision (ICCV)}, \bibinfo{organization}{IEEE}. pp.
  \bibinfo{pages}{5161--5169}.
%Type = Article
\bibitem[{Yang et~al.(2018)Yang, Cao, Wang, Zhang and Zou}]{yang2018counting}
\bibinfo{author}{Yang, B.}, \bibinfo{author}{Cao, J.}, \bibinfo{author}{Wang,
  N.}, \bibinfo{author}{Zhang, Y.}, \bibinfo{author}{Zou, L.},
  \bibinfo{year}{2018}.
\newblock \bibinfo{title}{Counting challenging crowds robustly using a
  multi-column multi-task convolutional neural network}.
\newblock \bibinfo{journal}{Signal Processing: Image Communication}
  \bibinfo{volume}{64}, \bibinfo{pages}{118--129}.
%Type = Inproceedings
\bibitem[{Yu and Koltun(2016)}]{yu2015multi}
\bibinfo{author}{Yu, F.}, \bibinfo{author}{Koltun, V.}, \bibinfo{year}{2016}.
\newblock \bibinfo{title}{Multi-scale context aggregation by dilated
  convolutions}, in: \bibinfo{booktitle}{ICLR}.
%Type = Inproceedings
\bibitem[{Zeng and Ma(2010)}]{zeng2010robust}
\bibinfo{author}{Zeng, C.}, \bibinfo{author}{Ma, H.}, \bibinfo{year}{2010}.
\newblock \bibinfo{title}{Robust head-shoulder detection by pca-based
  multilevel hog-lbp detector for people counting}, in:
  \bibinfo{booktitle}{2010 20th International Conference on Pattern
  Recognition}, \bibinfo{organization}{IEEE}. pp. \bibinfo{pages}{2069--2072}.
%Type = Article
\bibitem[{Zhan et~al.(2008)Zhan, Monekosso, Remagnino, Velastin and
  Xu}]{zhan2008crowd}
\bibinfo{author}{Zhan, B.}, \bibinfo{author}{Monekosso, D.N.},
  \bibinfo{author}{Remagnino, P.}, \bibinfo{author}{Velastin, S.A.},
  \bibinfo{author}{Xu, L.Q.}, \bibinfo{year}{2008}.
\newblock \bibinfo{title}{Crowd analysis: a survey}.
\newblock \bibinfo{journal}{Machine Vision and Applications}
  \bibinfo{volume}{19}, \bibinfo{pages}{345--357}.
%Type = Inproceedings
\bibitem[{Zhang et~al.(2015)Zhang, Li, Wang and Yang}]{zhang2015cross}
\bibinfo{author}{Zhang, C.}, \bibinfo{author}{Li, H.}, \bibinfo{author}{Wang,
  X.}, \bibinfo{author}{Yang, X.}, \bibinfo{year}{2015}.
\newblock \bibinfo{title}{Cross-scene crowd counting via deep convolutional
  neural networks}, in: \bibinfo{booktitle}{Proceedings of the IEEE conference
  on computer vision and pattern recognition}, pp. \bibinfo{pages}{833--841}.
%Type = Article
\bibitem[{Zhang et~al.(2019)Zhang, Zhou, Chang and Kot}]{zhang2019multi}
\bibinfo{author}{Zhang, Y.}, \bibinfo{author}{Zhou, C.},
  \bibinfo{author}{Chang, F.}, \bibinfo{author}{Kot, A.C.},
  \bibinfo{year}{2019}.
\newblock \bibinfo{title}{Multi-resolution attention convolutional neural
  network for crowd counting}.
\newblock \bibinfo{journal}{Neurocomputing} \bibinfo{volume}{329},
  \bibinfo{pages}{144--152}.
%Type = Inproceedings
\bibitem[{Zhang et~al.(2016)Zhang, Zhou, Chen, Gao and Ma}]{zhang2016single}
\bibinfo{author}{Zhang, Y.}, \bibinfo{author}{Zhou, D.}, \bibinfo{author}{Chen,
  S.}, \bibinfo{author}{Gao, S.}, \bibinfo{author}{Ma, Y.},
  \bibinfo{year}{2016}.
\newblock \bibinfo{title}{Single-image crowd counting via multi-column
  convolutional neural network}, in: \bibinfo{booktitle}{Proceedings of the
  IEEE conference on computer vision and pattern recognition}, pp.
  \bibinfo{pages}{589--597}.
%Type = Inproceedings
\bibitem[{Zhou et~al.(2016)Zhou, Khosla, Lapedriza, Oliva and
  Torralba}]{zhou2016learning}
\bibinfo{author}{Zhou, B.}, \bibinfo{author}{Khosla, A.},
  \bibinfo{author}{Lapedriza, A.}, \bibinfo{author}{Oliva, A.},
  \bibinfo{author}{Torralba, A.}, \bibinfo{year}{2016}.
\newblock \bibinfo{title}{Learning deep features for discriminative
  localization}, in: \bibinfo{booktitle}{Proceedings of the IEEE conference on
  computer vision and pattern recognition}, pp. \bibinfo{pages}{2921--2929}.

\end{thebibliography}


\bio[width=25mm]{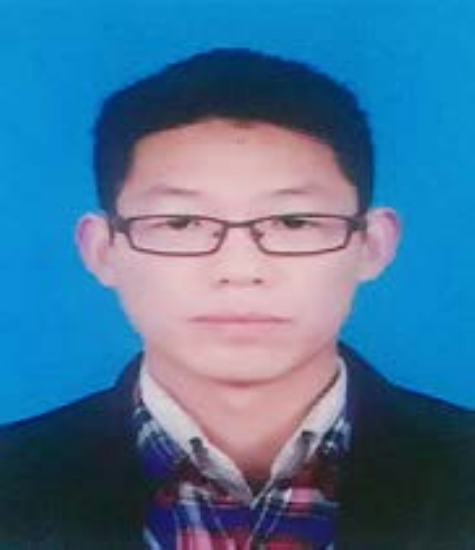}
\textbf{Jiwei Chen} is currently pursuing the Ph.D. degree in the Hefei Institutes of Physical Science, University of Science and Technology of China.
His research interests include crowd counting,
computer vision and machine learning.
\endbio

\vspace{80pt}

\bio[width=25mm]{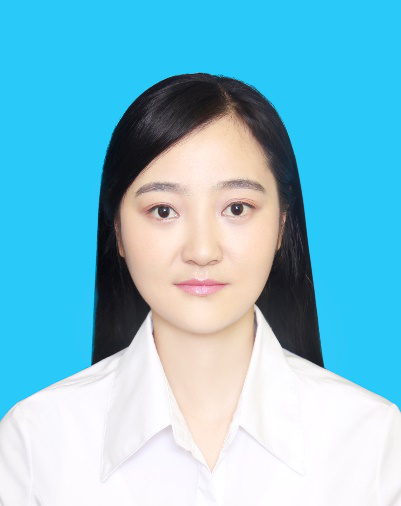}
\textbf{Wen Su} received Ph.D. degree in control science and engineering from University of Science and Technology of China in 2018 and B.E. degree in engineering from Automation Department, University of Science and Technology of China in 2013, respectively. Now, she works in virtual reality laboratory in Zhejiang Sci-Tech University. At present her research interests are image segmentation and depth scene understanding based on deep learning.
\endbio

\vspace{35pt}

%\bio[width=30mm,pos=l]{
\bio[width=25mm]{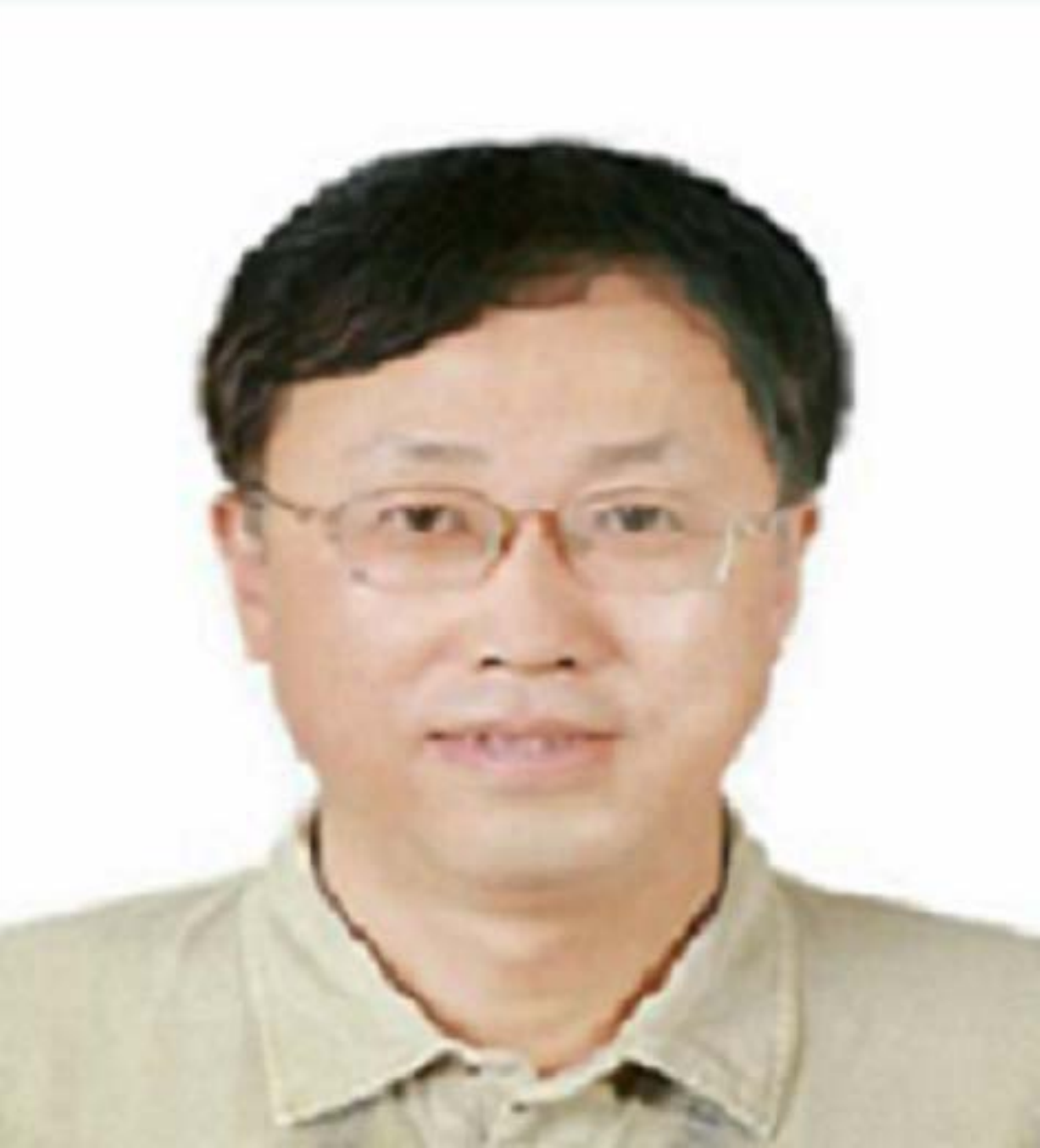}\textbf{Zengfu Wang} received the B.S. degree in electronic engineering from the University of Science and Technology of China in 1982 and the Ph.D. degree in control engineering from Osaka University, Japan, in 1992. He is currently a Professor with the Institute of Intelligent Machines, Chinese Academy of Sciences, and the Department of Automation, University of Science and Technology of China. He has published more than 300 journal articles and conference papers. His research interests include computer vision, human–computer interaction and intelligent robots. He received the Best Paper Award at the ACM International Conference on Multimedia 2009 and the IET Image Processing Premium Award 2017.
\endbio

\end{document}